\documentclass[10pt,twocolumn,letterpaper]{article}

\usepackage{cvpr}              
\usepackage{multirow}
\definecolor{cvprblue}{rgb}{0.21,0.49,0.74}
\usepackage[pagebackref,breaklinks,colorlinks,allcolors=cvprblue]{hyperref}
\def\paperID{14394} 
\def\confName{CVPR}
\def\confYear{2025}

\title{InPO: Inversion Preference Optimization with Reparametrized DDIM for Efficient Diffusion Model Alignment}

\author{Yunhong Lu$^1$ {\quad} Qichao Wang$^1$ {\quad} Hengyuan Cao$^1$ {\quad} Xierui Wang$^1$ {\quad} Xiaoyin Xu$^1$ {\quad} Min Zhang$^{1,2*}$ 
\\
$^1$ Zhejiang University{\quad}$^2$ Shanghai Institute for Advanced Study-Zhejiang University\\
{\tt\small \{yunhonglu,qichaowang,caohy,sherrywang,xiaoyinxu,min\_zhang\}@zju.edu.cn}
}

\newcounter{suppsection}
\renewcommand{\thesuppsection}{S\arabic{suppsection}}
\newcommand{\suppref}[1]{Supp.~\ref{#1}}
\newcommand{\suppsection}[1]{%
    \refstepcounter{suppsection}%
    \section*{\thesuppsection: #1}%
    \addcontentsline{toc}{section}{Supplementary Section \thesuppsection: #1}%
    \label{suppsec:\thesuppsection}
}

\begin{document}
\maketitle
\let\thefootnote\relax\footnote{\scriptsize{* Corresponding author}}
\begin{abstract}
\label{sec:abstract}

Without using explicit reward, direct preference optimization (DPO) employs paired human preference data to fine-tune generative models, a method that has garnered considerable attention in large language models (LLMs). However, exploration of aligning text-to-image (T2I) diffusion models with human preferences remains limited. In comparison to supervised fine-tuning, existing methods that align diffusion model suffer from low training efficiency and subpar generation quality due to the long Markov chain process and the intractability of the reverse process. To address these limitations, we introduce DDIM-InPO, an efficient method for direct preference alignment of diffusion models. Our approach conceptualizes diffusion model as a single-step generative model, allowing us to fine-tune the outputs of specific latent variables selectively. In order to accomplish this objective, we first assign implicit rewards to any latent variable directly via a reparameterization technique. Then we construct an Inversion technique to estimate appropriate latent variables for preference optimization. This modification process enables the diffusion model to only fine-tune the outputs of latent variables that have a strong correlation with the preference dataset. Experimental results indicate that our DDIM-InPO achieves state-of-the-art performance with just 400 steps of fine-tuning, surpassing all preference aligning baselines for T2I diffusion models in human preference evaluation tasks. All resources will be available at \url{https://github.com/JaydenLyh/InPO}.
\end{abstract}    
\section{Introduction}
\label{sec:intro}
In the domain of T2I generation, tailoring image outputs to reflect user preferences is still in an early stage of development. In contrast, large language models (LLMs) \cite{achiam2023gpt} have made notable progress in generating human-aligned text, following a two-stage process: first, pre-training on extensively collected large-scale datasets and then, aligning on smaller datasets that capture specific human needs. This alignment aims to produce outputs that meet human preferences while retaining model's learned capabilities. 
In recent years, T2I diffusion models \cite{rombach2022high,podell2023sdxl,Esser2024ScalingRF} have been at the forefront of image generation. Nevertheless, achieving alignment between model outputs and human preferences remains a significant challenge.

\begin{figure*}[t]
  \centering
   \includegraphics[width=1\linewidth]{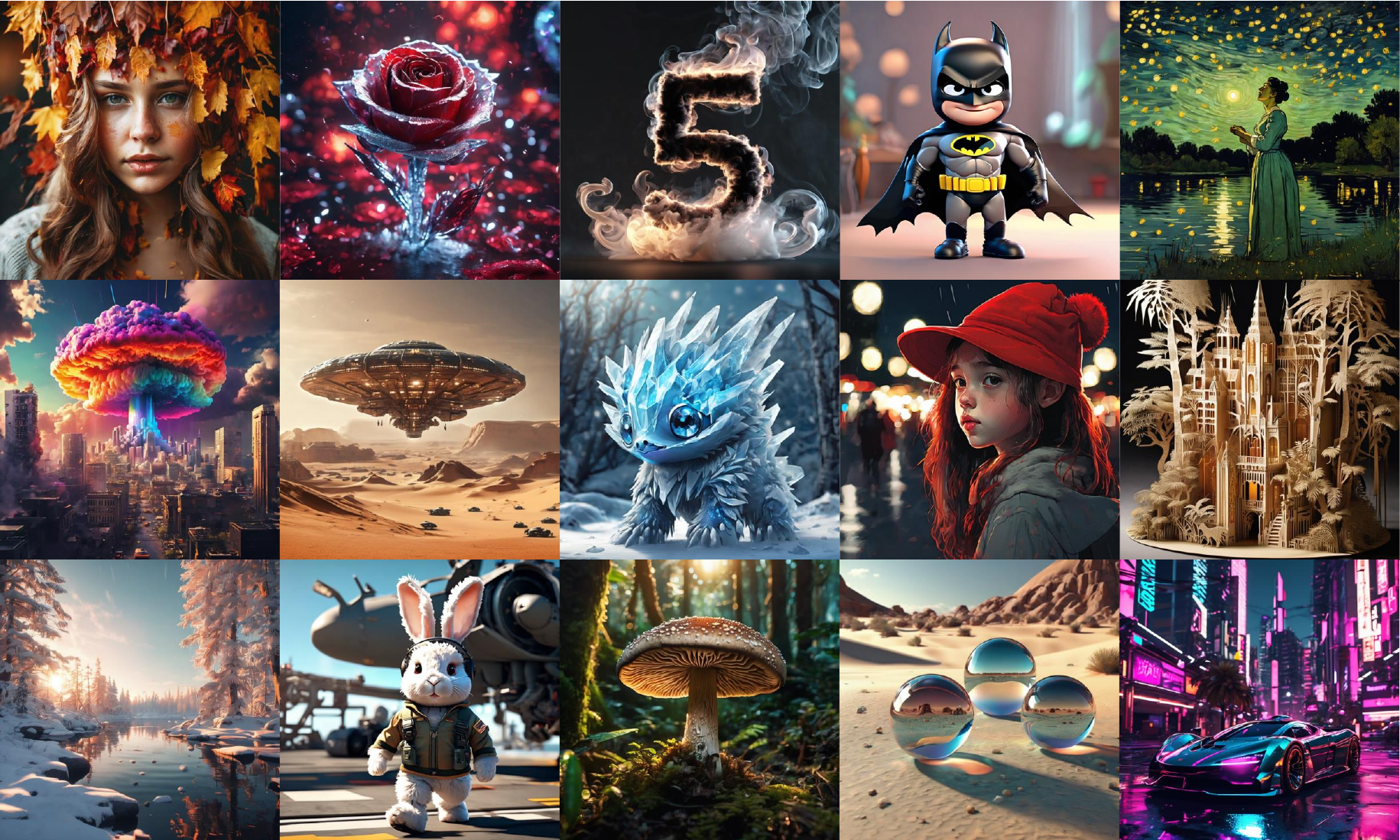}

   \caption{We develop DDIM-InPO, an efficient method to align diffusion models with human preference. It suffices to directly optimize the outputs of a small set of variables using human feedback data. This figure illustrates the results after 400 fine-tuning steps of SDXL-base-1.0 using our method, showing that the generated images exhibit strong visual appeal and align well with human preferences.}
   \label{fig:inpodemo}
\end{figure*}

Recently, researchers have developed several methods tailoring diffusion models to human preferences, yet these approaches generally lack effectiveness. 
Some methods adopt a two-stage training approach, where a reward function is first trained and then followed by updating the diffusion model using gradient-based back-propagation \cite{Prabhudesai2023AligningTD, Clark2023DirectlyFD}. These approaches tend to be unstable and are susceptible to reward leakage \cite{zhang2024large}. Alternatively, RL (reinforcement learning)-based methods model the denoising process as a Markov decision process, with online evaluation of generated images based on reward feedback \cite{Black2023TrainingDM}. However, their effectiveness is constrained by the length of the Markov chain. Other approaches such as~\cite{wallace2024diffusion,li2024aligning}, inspired by direct preference optimization (DPO) \cite{rafailov2024direct}, reduce the RL objective to an optimal policy training goal. Nevertheless, due to the influence of sparse rewards \cite{Guo2021EfficientQ, Yang2024ADR}, these methods' performance is limited, especially when handling out-of-distribution inputs.

To address these issues, we introduce DDIM-InPO, an efficient approach for direct preference optimization of diffusion models. In response to the challenge of the implict rewards allocation in the long-chain denoising process, our key idea is to treat the diffusion model as a single-step generation framework, only fine-tuning the outputs of latent variables directly linked to the target image. To achieve this, we first link the latent variables to the target image space in a single step, using a reparameterization method. It allows us to allocate implicit rewards to latent variables at any timestep directly. Secondly, to mitigate the effects of sparse rewards, we use an Inversion technique in the target image space to identify images that are highly correlated with the target and obtain the corresponding latent variables. Finally, we focus on optimizing the outputs of this limited set of latent variables and thus achieve rapid computations, enabling high-quality optimization of the diffusion model. Our main contributions are three-folds:
\begin{itemize}
    \item We propose a novel DPO implicit rewards allocation method for the diffusion model, treating the diffusion model as a single-step generative model.
    \item We introduce DDIM-InPO, an efficient framework for direct preference optimization of diffusion models. By applying an Inversion technique in the target image space, we estimate images highly correlated with the preference dataset and derive the corresponding latent variables sufficient for fine-tuning. 
    \item In our experiments, we benchmark our method against state-of-the-art fine-tuning approaches for human preference, demonstrating enhanced training efficiency (refer to \cref{fig:speedsd15}) and improved generation quality (examples shown in \cref{fig:inpodemo} and \cref{fig:inpoadvantage1}).
\end{itemize}
\section{Related Works}
\label{sec:relatedworks}
\paragraph{Text-to-Image Generative Models}
T2I generation models aim to produce high-quality, high-fidelity images from a given text prompt. Additionally, these models have contributed to advancements in related fields such as image editing \cite{Huang2024DiffusionMI}, video generation \cite{Chen2023VideoCrafter1OD}, and 3D modeling \cite{Poole2022DreamFusionTU}. 
Among T2I generative models \cite{Reed2016GenerativeAT, Dao2023FlowMI, Ramesh2021ZeroShotTG, Ramesh2022HierarchicalTI, Yu2022ScalingAM}, diffusion models have recently gained significant attention due to their open-source nature and superior generation performance. However, high-quality diffusion models are typically trained on large, noisy datasets, which often leads to deviations from human preferences, an area that still requires further refinement.

\paragraph{Diffusion Models Alignment}
In diffusion models, supervised fine-tuning (SFT) is commonly employed to align them with human preferences. Inspired by RL-based fine-tuning methods used in LLMs \cite{Schulman2017ProximalPO, Azar2023AGT, Hong2024ORPOMP, Ethayarajh2024KTOMA, Song2023PreferenceRO} (refer to \suppref{sec:supprelatedworks} for related LLMs alignment), various reward models such as PickScore \cite{kirstain2023pick} and HPSv2 \cite{Wu2023HumanPS} have been developed to simulate human preferences. With reward models, approaches like DRaFT \cite{Clark2023DirectlyFD} and AlignProp \cite{Prabhudesai2023AligningTD} use differentiable rewards and back-propagate through the diffusion sampling process, which demands substantial memory and would result in reward leakage. DPOK \cite{Fan2023DPOKRL} and DDPO \cite{Black2023TrainingDM} model the sampling process of diffusion models as a Markov decision process and apply policy gradient methods for fine-tuning. Diffusion-DPO \cite{wallace2024diffusion} and D3PO \cite{Yang2023UsingHF} adopt direct preference optimization by leveraging the denoising steps of diffusion models to optimize model parameters at each step. In the DPO framework, DenseReward \cite{Yang2024ADR} allocates higher weights to the initial steps. Going further, Diffusion-KTO \cite{li2024aligning} requires only binary feedback signals for each image to fine-tune the model. However, all these methods are limited to expanding the decision space by incorporating the reward function over all denoising steps, which results in sparse rewards and inefficient training.
\section{Preliminaries}
\label{sec:pre}
\subsection{Diffusion Models}
Diffusion models involve a forward process $\{q(\boldsymbol{x}_{t})\}_{t\in[0,T]}$ that incrementally adds noise to data $\boldsymbol{x}_{0} \sim q(\boldsymbol{x}_{0})$ and a learned reverse process $\{p(\boldsymbol{x}_{t})\}_{t\in[0,T]}$ that aims to denoise the data. The forward process is defined as $q(\boldsymbol{x}_{t}|\boldsymbol{x}_{0}):=\mathcal{N}(\sqrt{\alpha_{t}}\boldsymbol{x}_{0},(1-\alpha_{t})\mathbf{I})$ and $q(\boldsymbol{x}_{t}):= \int q(\boldsymbol{x}_{t}|\boldsymbol{x}_{0})q(\boldsymbol{x}_{0}) \mathrm{d}\boldsymbol{x}_{0}$, where $\alpha_{t}$ is a noise schedule (as defined in \cite{song2020denoising}). Starting from $p(\boldsymbol{x}_{T}):=\mathcal{N}(\mathbf{0},\mathbf{I})$, the reverse process is defined by a parameterized denoiser $\boldsymbol{\epsilon}_{\theta}^{t}(\boldsymbol{x}_{t})$, which aims to predict noise added to $\boldsymbol{x}_{0}$. The denoiser is optimized by minimizing:
\begin{equation}
  \mathcal{L}_{\mathrm{DM}}:=\mathbb{E}_{x_{0},t,\boldsymbol{\epsilon}}[w(t)\left\|\boldsymbol{\epsilon}_{\theta}^{t}(\sqrt{\alpha_{t}}\boldsymbol{x}_{0}+\sqrt{1-\alpha_{t}}\boldsymbol{\epsilon})-\boldsymbol{\epsilon}\right\|^{2}_{2}]
  \label{eq:diffusion_loss}
\end{equation}
where $\boldsymbol x_{0}\sim q(\boldsymbol x_{0}),t\sim \mathcal{U}(0,T),\boldsymbol{\epsilon} \sim \mathcal{N}(\mathbf{0},\mathbf{I})$ and $w(t)$ is a pre-specified weight function.

Denoising Diffusion Implicit Models (DDIM) generate samples by deterministically reversing the diffusion process. By reparameterizing noisy image $\Bar{\boldsymbol{x}}_{t}=\boldsymbol{x}_{t}/\sqrt{\alpha_{t}}$, the denoising process can be viewed as an ODE:
\begin{equation}
\frac{\mathrm{d}\Bar{\boldsymbol{x}}_{t}}{\mathrm{d}t}=\boldsymbol{\epsilon}^{t}_{\theta}\left(\frac{\Bar{\boldsymbol{x}}_{t}}{\sqrt{\sigma_{t}^{2}+1}}\right)\frac{\mathrm{d}\sigma_{t}}{\mathrm{d}t}
\label{eq:ddim_ode}
\end{equation}
where $\sigma_{t}=\sqrt{1-\alpha_{t}}/\sqrt{\alpha_{t}}$ and $\Bar{\boldsymbol{x}}_{T}$ is sampled from Gaussian distribution. 
The DDIM ODE can also be integrated in reverse to estimate $\boldsymbol{x}_{t}$ at any timestep $t$ starting from a clean image $\boldsymbol{x}_{0}$, a process known as DDIM inversion \cite{mokady2023null}.
\subsection{Human Preference Optimization}
Preference optimization involves fine-tuning generative models to produce samples more closely align with human preferences. Assuming that human preferences are represented by an implicit reward function $r^{*}$, there is a dataset $\mathcal{D}=\{(\boldsymbol{x}^{w}_{0},\boldsymbol{x}^{l}_{0},\boldsymbol{c})\}$, where $\boldsymbol{c}$ denotes the condition, and $\boldsymbol{x}^{w}_{0}$ and $\boldsymbol{x}^{l}_{0}$ represent the winner and loser samples, respectively.
\paragraph{RLHF}
RLHF employs maximum likelihood estimation to fit the Bradley-Terry (BT) model on the dataset $\mathcal{D}$, thereby obtaining an explicit parameterized reward model $r_{\phi}$.
\begin{equation}
\mathcal{L}_{\mathrm{BT}}(\phi) := -\mathbb{E}_{(\boldsymbol{x}^{w}_{0},\boldsymbol{x}^{l}_{0},\boldsymbol{c})\sim \mathcal{D}}[\log \sigma(r_{\phi}(\boldsymbol{x}^{w}_{0},\boldsymbol{c})-r_{\phi}(\boldsymbol{x}^{l}_{0},\boldsymbol{c}))]
\label{eq:btmodel}
\end{equation}
With the reward, the generative distribution $p_{\theta}$ can then be optimized as a policy using RL that maximizes reward feedback, with an regularization term for the KL-divergence from the reference distribution $p_{\mathrm{ref}}$:
\begin{equation}
    \max_{p_{\theta}} \mathbb{E}_{\substack{\boldsymbol{c}\sim \mathcal{D}_{\boldsymbol{c}}\\ \boldsymbol{x}_{0}\sim p_{\theta}(\boldsymbol{x}_{0}|\boldsymbol{c})}}[r(\boldsymbol{x}_{0},\boldsymbol{c})]-\beta \mathbb{D}_{\mathrm{KL}}[p_{\theta}(\boldsymbol{x}_{0}|\boldsymbol{c})||p_{\mathrm{ref}}(\boldsymbol{x}_{0}|\boldsymbol{c})]
    \label{eq:rlhf}
\end{equation}
where $\beta$ is a hyperparameter that controls regularization.
\paragraph{Direct Preference Optimization}
Direct Preference Optimization (DPO) considers the unique global optimal policy $p_{\theta}^{*}$ under the given reward function $r$ in \cref{eq:rlhf}:
\begin{equation}
    p_{\theta}^{*}(\boldsymbol{x}_{0}|\boldsymbol{c}) = \frac{p_{\mathrm{ref}}(\boldsymbol{x}_{0}|\boldsymbol{c})\exp(r(\boldsymbol{x}_{0},\boldsymbol{c})/\beta)}{\sum_{\boldsymbol{x}_{0}}p_{\mathrm{ref}}(\boldsymbol{x}_{0}|\boldsymbol{c})\exp(r(\boldsymbol{x}_{0},\boldsymbol{c})/\beta)}
    \label{eq:optimalpolicy}
\end{equation}
Then, the reward function $r(\boldsymbol{x}_{0},\boldsymbol{c})$ is rewritten according to \cref{eq:optimalpolicy} and inserted into the BT model from \cref{eq:btmodel}, resulting in the DPO objective function:
\begin{footnotesize}
\begin{equation}
    \mathcal{L}_{\mathrm{DPO}} := -\mathbb{E}_{\substack{(\boldsymbol{x}^{w}_{0},\boldsymbol{x}^{l}_{0}\\ ,\boldsymbol{c})\sim \mathcal{D}}}\log \sigma \left(\beta \log \frac{p_{\theta}(\boldsymbol{x}_{0}^{w}|\boldsymbol{c})}{p_{\mathrm{ref}}(\boldsymbol{x}_{0}^{w}|\boldsymbol{c})}-\beta \log \frac{p_{\theta}(\boldsymbol{x}_{0}^{l}|\boldsymbol{c})}{p_{\mathrm{ref}}(\boldsymbol{x}_{0}^{l}|\boldsymbol{c})}\right)
    \label{eq:originaldpo}
\end{equation}
\end{footnotesize}Based on \cref{eq:rlhf,eq:originaldpo}, Wallace \etal \cite{wallace2024diffusion} redistribute rewards $r(\boldsymbol{x}_{0},c)$ across all possible diffusion paths $p_{\theta}(\boldsymbol{x}_{1:T}|\boldsymbol{x}_{0},\boldsymbol{c})$ and minimize the KL-divergence of the joint probability density $\mathbb{D}_{\mathrm{KL}}(p_{\theta}(\boldsymbol{x}_{0:T}|\boldsymbol{c})||p_{\mathrm{ref}}(\boldsymbol{x}_{0:T}|\boldsymbol{c}))$, thus designing the DPO objective for diffusion models:
\begin{equation}
\begin{split}
    &\mathcal{L}_{\mathrm{DPO-Diffusion}}:= -\mathbb{E}_{(\boldsymbol{x}^{w}_{0},\boldsymbol{x}^{l}_{0}, \boldsymbol{c})\sim \mathcal{D}}\log \sigma 
    \\
    &\left(\beta\mathbb{E}_{\substack{\boldsymbol{x}_{1:T}^{w} \sim p^{\boldsymbol{c}}_{\theta}(\boldsymbol{x}_{1:T}^{w}|\boldsymbol{x}_{0}^{w})\\\boldsymbol{x}_{1:T}^{l} \sim p^{\boldsymbol{c}}_{\theta}(\boldsymbol{x}_{1:T}^{l}|\boldsymbol{x}_{0}^{l})}}\left[ \log \frac{p^{\boldsymbol{c}}_{\theta}(\boldsymbol{x}_{0:T}^{w})}{p^{\boldsymbol{c}}_{\mathrm{ref}}(\boldsymbol{x}_{0:T}^{w})}-\log \frac{p^{\boldsymbol{c}}_{\theta}(\boldsymbol{x}_{0:T}^{l})}{p^{\boldsymbol{c}}_{\mathrm{ref}}(\boldsymbol{x}_{0:T}^{l})}\right]\right)
\end{split}
\label{eq:diffusiondpo}
\end{equation}
where $p^{\boldsymbol{c}}_{\theta}(\cdot)$ represents $p_{\theta}(\cdot| \boldsymbol{c})$ for compactness. Without the need for RL algorithms, the DPO-based method directly optimizes the optimal distribution $p_{\theta}$.
\section{Inversion Preference Optimization}
\label{sec:inpo}
Given a fixed dataset  $\mathcal{D}=\{(\boldsymbol{x}^{w}_{0},\boldsymbol{x}^{l}_{0},\boldsymbol{c})\}$, where each sample includes a prompt $\boldsymbol{c}$ and an image pair generated by reference model $p_{\mathrm{ref}}$ labeled as $\boldsymbol{x}^{w}_{0} \succ \boldsymbol{x}^{l}_{0}$, our goal is to train a model $p_{\theta}$ that aligns with human preferences. 
The primary challenge lies in how to match the parameterized distribution $p_{\theta}^{\boldsymbol{c}}(\boldsymbol{x}_{0})$ according to \cref{eq:originaldpo}. Our \textbf{insight} is that the diffusion model can be conceptualized as a timestep-aware single-step generative model $p_{\theta}^{\boldsymbol{c}}(\boldsymbol{x}_{0}|\boldsymbol{x}_{t})$. Specifically, there exists some latent variables $\boldsymbol{x}_{t}$ that can generate $\boldsymbol{x}_{0}$ in a single step for any timestep $t$, which can be written mathematically as:
$p_{\theta}^{\boldsymbol{c}}(\boldsymbol{x}_{0}) = \int p_{\theta}^{\boldsymbol{c}}(\boldsymbol{x}_{0}|\boldsymbol{x}_{t}) p_{\theta}^{\boldsymbol{c}}(\boldsymbol{x}_{t}) \mathrm{d}\boldsymbol{x}_{t}$. Thus, the \textbf{first step} is to determine the relationship between $\boldsymbol{x}_{t}$ and $\boldsymbol{x}_{0}$ in \cref{subsec:ddimdpo}, and the \textbf{second step} is to identify appropriate latent variables $\boldsymbol{x}_{t}$ for any timestep in \cref{subsec:poi}, within the DPO framework. 
\subsection{Reparametrized DDIM for DPO}
\label{subsec:ddimdpo}
We establish a relationship between $\boldsymbol{x}_{t}$ and $\boldsymbol{x}_{0}$ to be used within the DPO framework. 
Instead of viewing DDIM as a denoising process within the noisy image $\boldsymbol{x}_{t}$ space, we reparameterize it with \textquoteleft initial' variable:
\begin{equation}
    \boldsymbol{x}_{0}(t):= \Bar{\boldsymbol{x}}_{t}-\sigma_{t}\epsilon_{\theta}^{t}(\boldsymbol{x}_{t},\boldsymbol{c})
    \label{eq:initialvariable}
\end{equation}
where noisy image $\Bar{\boldsymbol{x}}_{t} = \boldsymbol{x}_{t}/\sqrt{\alpha_{t}}$. In this context, $\boldsymbol{x}_{0}(t)$ is the image obtained after the single-step denoising at timestep $t$ and is in $\boldsymbol{x}_{0}$ space. Therefore, $p^{\boldsymbol{c}}_{\theta}(\boldsymbol{x}_{0},\boldsymbol{x}_{t})$ is well-defined, allowing us to present a formal definition of \textquoteleft initial'  reward $ r(\boldsymbol{x}_{0},\boldsymbol{c})$ with the \textquoteleft joint' reward $r^{\boldsymbol{c}}_{t}(\boldsymbol{x}_{0},\boldsymbol{x}_{t})$:
\begin{equation}
    r(\boldsymbol{x}_{0},\boldsymbol{c}) = \mathbb{E}_{p^{\boldsymbol{c}}_{\theta}(\boldsymbol{x}_{t}|\boldsymbol{x}_{0})}[r^{\boldsymbol{c}}_{t}(\boldsymbol{x}_{0},\boldsymbol{x}_{t})]
    \label{eq:rewardfunction}
\end{equation}
Similar to \cite{wallace2024diffusion}, we minimize the joint KL divergence $\mathbb{D}_{\mathrm{KL}}[p^{\boldsymbol{c}}_{\theta}(\boldsymbol{x}_{0},\boldsymbol{x}_{t})||p^{\boldsymbol{c}}_{\mathrm{ref}}(\boldsymbol{x}_{0},\boldsymbol{x}_{t})]$ to serve as an upper bound for the KL-divergence term in \cref{eq:rlhf}. 
With \cref{eq:rewardfunction} and  KL-divergence bound, we can reformulate \cref{eq:rlhf} into the objective below:
\begin{equation}
    \max_{p^{\boldsymbol{c}}_{\theta}} \mathbb{E}_{\substack{t,(\boldsymbol{x}_{0},\boldsymbol{x}_{t}) \sim \\ p^{\boldsymbol{c}}_{\theta}(\boldsymbol{x}_{0},\boldsymbol{x}_{t})}}[r_{t}^{\boldsymbol{c}}(\boldsymbol{x}_{0},\boldsymbol{x}_{t})]-\beta \mathbb{D}_{\mathrm{KL}}[p^{\boldsymbol{c}}_{\theta}(\boldsymbol{x}_{0},\boldsymbol{x}_{t})||p^{\boldsymbol{c}}_{\mathrm{ref}}(\boldsymbol{x}_{0},\boldsymbol{x}_{t})]
    \label{eq:jointrlhf}
\end{equation}
In this context, we consider all timesteps.
This objective enables the effective allocation of $\boldsymbol{x}_{0}$'s reward across any timestep within the model, enhancing training efficiency. 
In a manner analogous to \cref{eq:btmodel} through \cref{eq:originaldpo}, we derive the following optimization objective:
\begin{equation}
\begin{split}
    &\mathcal{L}(\theta):= -\mathbb{E}_{t,(\boldsymbol{x}^{w}_{0},\boldsymbol{x}^{l}_{0},\boldsymbol{c})\sim \mathcal{D}}\log \sigma 
    \\
    &\left(\beta\mathbb{E}_{\substack{\boldsymbol{x}_{t}^{w} \sim p^{\boldsymbol{c}}_{\theta}(\boldsymbol{x}_{t}^{w}|\boldsymbol{x}_{0}^{w})\\\boldsymbol{x}_{t}^{l} \sim p^{\boldsymbol{c}}_{\theta}(\boldsymbol{x}_{t}^{l}|\boldsymbol{x}_{0}^{l})}}\left[ \log \frac{p^{\boldsymbol{c}}_{\theta}(\boldsymbol{x}_{0}^{w},\boldsymbol{x}_{t}^{w})}{p^{\boldsymbol{c}}_{\mathrm{ref}}(\boldsymbol{x}_{0}^{w},\boldsymbol{x}_{t}^{w})}-\log \frac{p^{\boldsymbol{c}}_{\theta}(\boldsymbol{x}_{0}^{l},\boldsymbol{x}_{t}^{l})}{p^{\boldsymbol{c}}_{\mathrm{ref}}(\boldsymbol{x}_{0}^{l},\boldsymbol{x}_{t}^{l})}\right]\right)
\end{split}    
    \label{eq:dpoddim}
\end{equation}
(refer to \suppref{sec:suppprimaryderivation} for details). To optimize \cref{eq:dpoddim}, we need to sample $\boldsymbol{x}_{t}\sim p^{\boldsymbol{c}}_{\theta}(\boldsymbol{x}_{t}|\boldsymbol{x}_{0})$, where $\boldsymbol{x}_{t}$ should generate an approximation of $\boldsymbol{x}_{0}$ in a single step. 
\begin{figure}[t]
  \centering
   \includegraphics[width=1\linewidth]{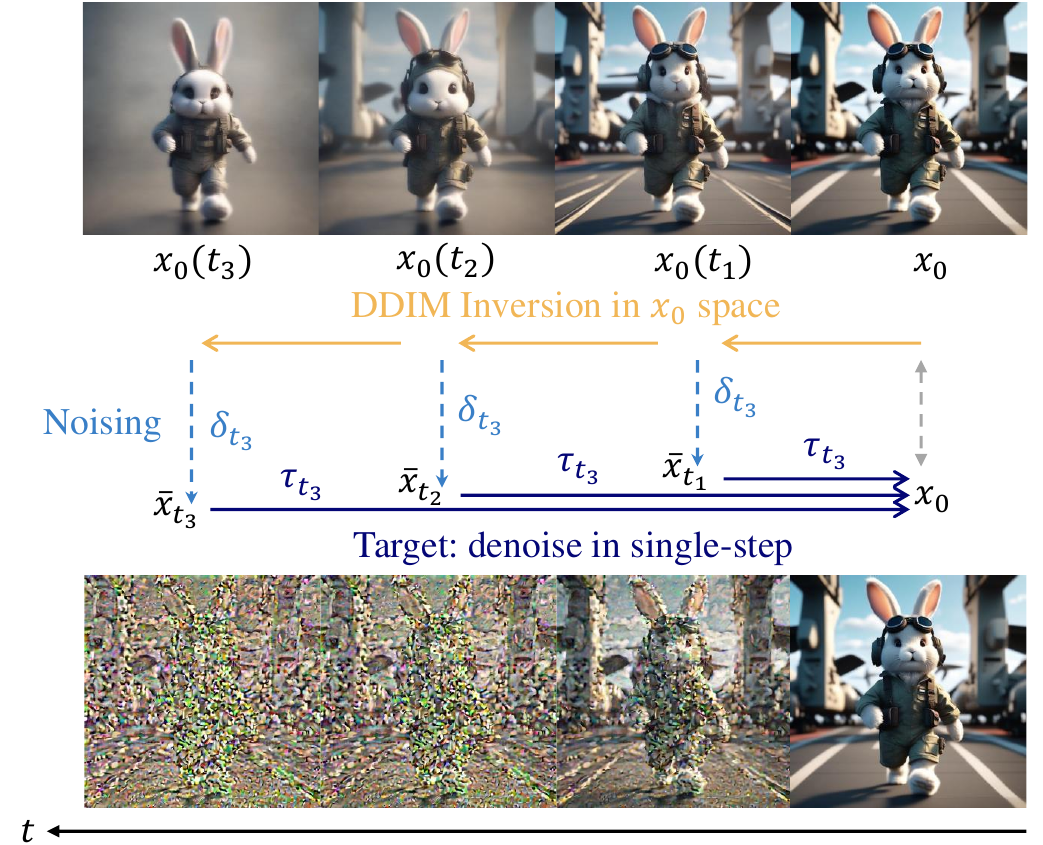}

   \caption{Illustration of Inversion for Preference Optimization. }
   \label{fig:pipeline}
\end{figure}
\subsection{Preference Optimization via Inversion}
\label{subsec:poi}
\textit{So how can we determine whether $\boldsymbol{x}_{t}$ can generate an approximation of $\boldsymbol{x}_{0}$ in a single step?} 
Given $\boldsymbol{x}_{0}(t)$, it remains unclear how to calculate $\boldsymbol{x}_{t}$. 
From \cref{eq:initialvariable}, $\boldsymbol{x}_{t}$ should satisfy the following nonlinear equation:
\begin{equation}
    \boldsymbol{x}_{t} = \sqrt{\alpha_{t}}\boldsymbol{x}_{0}(t)+\sqrt{1-\alpha_{t}}\epsilon_{\theta}^{t}(\boldsymbol{x}_{t},\boldsymbol{c})
    \label{eq:nonlinear}
\end{equation}
and we can rewrite \cref{eq:nonlinear} with respect to noise:
\begin{equation}
    \epsilon=\epsilon_{\theta}^{t}(\sqrt{\alpha_{t}}\boldsymbol{x}_{0}(t)+\sqrt{1-\alpha_{t}}\epsilon,\boldsymbol{c})
    \label{eq:epsilonnonlinear}
\end{equation}
where $\epsilon=(\boldsymbol{x}_{t}-\sqrt{\alpha_{t}}\boldsymbol{x}_{0}(t))/\sqrt{1-\alpha_{t}}$. Thus, we can define $\delta_{t}(\boldsymbol{x}_{0}(t))=\epsilon$ as the solution when given $\boldsymbol{x}_{0}(t)$.
In this context, $\boldsymbol{x}_{0}(t)$ serves as a reliable approximation of $\boldsymbol{x}_{0}$ at timestep $t$, that is, $\boldsymbol{x}_{0}(t) \approx \boldsymbol{x}_{0}$.
Additionally, we rewrite \cref{eq:ddim_ode} by applying the definition of $\boldsymbol{x}_{0}(t)$ (\cref{eq:initialvariable}) by employing some algebra:
\begin{equation}
    \frac{\mathrm{d}\boldsymbol{x}_{0}(t)}{\mathrm{d}t}=-\sigma_{t}\frac{\mathrm{d}}{\mathrm{d}t}\epsilon_{\theta}^{t}(\sqrt{\alpha_{t}}\boldsymbol{x}_{0}(t)+\sqrt{1-\alpha_{t}}\delta_{t}(\boldsymbol{x}_{0}(t)),\boldsymbol{c})
    \label{eq:initialddim}
\end{equation}
Inspired by the work of \cite{lukoianov2024score, mokady2023null}, we suggest to derive $\delta_{t}(\boldsymbol{x}_{0}(t))$ by DDIM inversion (\textit{It should be noted that $\delta_{t}(\boldsymbol{x}_{0}(t))$ can be estimated using various methods, including but not limited to, inversion.} \suppref{sec:suppdelta} includes more choices.) and obtain $\boldsymbol{x}_{0}(t)$ by inverting \cref{eq:initialddim} consequently. 
To clarify the process,  we consider the consecutive time points $t$ and $t+\kappa>t$, and the discretized form of inversion is given as follows :
\begin{equation}
    \begin{split}
        &\boldsymbol{x}_{0}(t+\kappa) = \boldsymbol{x}_{0}(t)- \sigma_{t+\kappa}[\epsilon_{\theta}^{t+\kappa}(\sqrt{\alpha_{t+\kappa}}\boldsymbol{x}_{0}(t)\\
        &+\sqrt{1-\alpha_{t+\kappa}}\delta_{t}(\boldsymbol{x}_{0}(t)))-\delta_{t}(\boldsymbol{x}_{0}(t))]
    \end{split}
    \label{eq:intialddiminversion}
\end{equation}
\begin{figure*}[ht]
  \centering
   \includegraphics[width=1\linewidth]{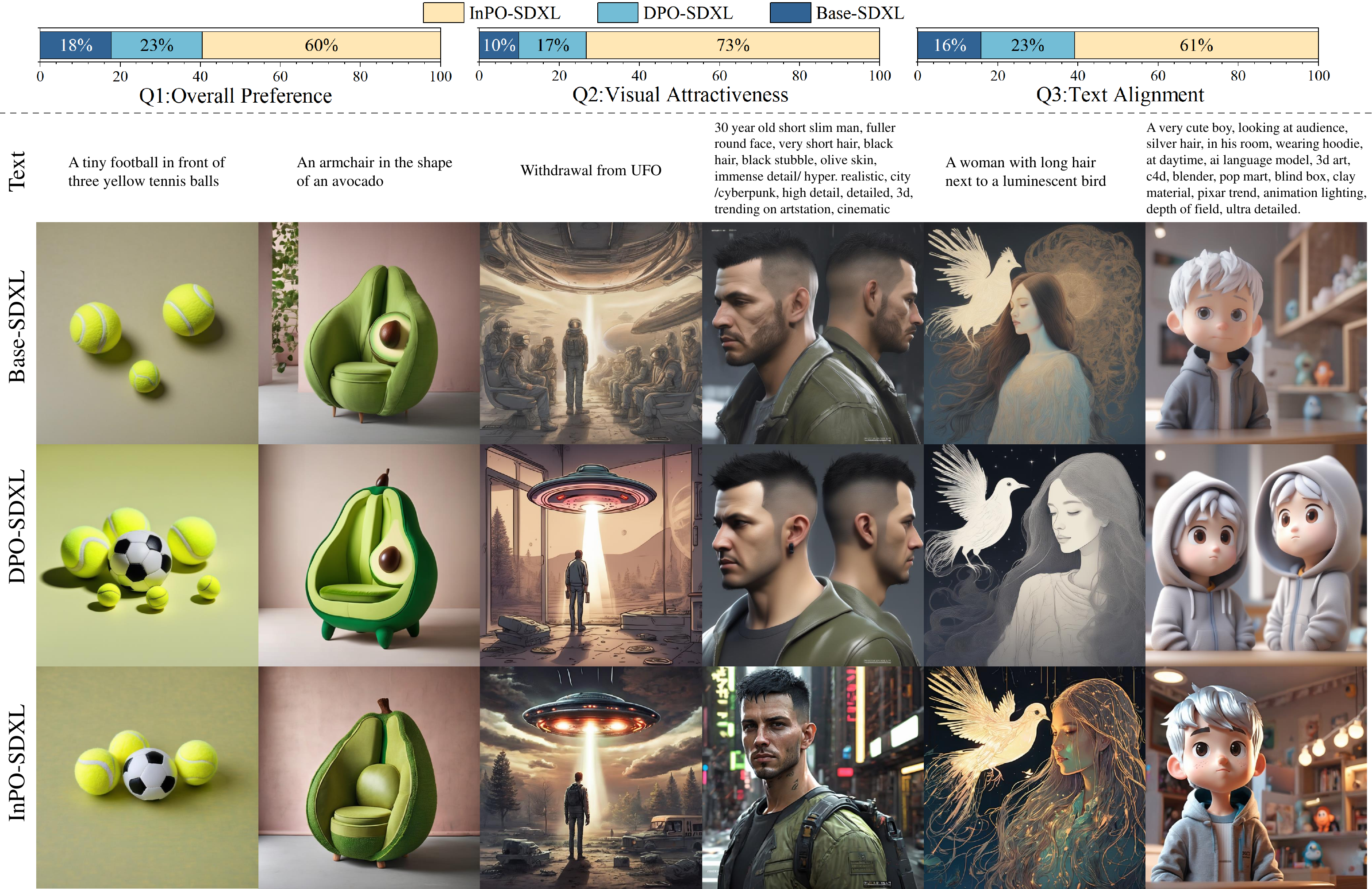}

   \caption{Top, in human evaluations, InPO-SDXL shows a marked improvement over both DPO-SDXL and Base-SDXL. Bottom, qualitative comparisons among baselines. InPO-SDXL achieves superior prompt alignment and produces images of higher quality.}
   \label{fig:inpoadvantage1}
\end{figure*}According to \cref{eq:nonlinear}, $\boldsymbol{x}_{t}$ is well-defined as follows, and so is $p^{\boldsymbol{c}}_{\theta}(\boldsymbol{x}_{t}|\boldsymbol{x}_{0})$, ensuring that \cref{eq:dpoddim} can be optimized.
\begin{equation}
    \boldsymbol{x}_{t} = \sqrt{\alpha_{t}}\boldsymbol{x}_{0}(t)+\sqrt{1-\alpha_{t}}\delta_{t}(\boldsymbol{x}_{0}(t))
    \label{eq:xtsolution}
\end{equation}
For the sake of simplicity, we approximate $p^{\boldsymbol{c}}_{\mathrm{ref}}(\boldsymbol{x}_{0},\boldsymbol{x}_{t})$ with $p^{\boldsymbol{c}}_{\mathrm{ref}}(\boldsymbol{x}_{0}|\boldsymbol{x}_{t})p^{\boldsymbol{c}}_{\theta}(\boldsymbol{x}_{t})$. Using Jensen’s inequality and some algebra (details in \suppref{sec:suppprimaryderivation}), we approximate \cref{eq:dpoddim} as:
\begin{equation}
\begin{split}
    \mathcal{L}(\theta)=& -\mathbb{E}_{\substack{(\boldsymbol{x}^{w}_{0},\boldsymbol{x}^{l}_{0}, \boldsymbol{c})\sim \mathcal{D},t \sim \mathcal{U}(0,T), \\ \boldsymbol{x}_{t}^{w}  \sim p^{\boldsymbol{c}}_{\theta}(\boldsymbol{x}_{t}^{w}|\boldsymbol{x}_{0}^{w}), \boldsymbol{x}_{t}^{l} \sim p^{\boldsymbol{c}}_{\theta}(\boldsymbol{x}_{t}^{l}|\boldsymbol{x}_{0}^{l})}} \log \sigma (-\beta(\\ 
    &+\mathbb{D}_{\mathrm{KL}}(q^{\boldsymbol{c}}(\boldsymbol{x}^{w}_{0}|\boldsymbol{x}^{w}_{t},\boldsymbol{x}^{w}_{0}(t))\lVert p^{\boldsymbol{c}}_{\theta}(\boldsymbol{x}^{w}_{0}|\boldsymbol{x}^{w}_{t}))\\
    &-\mathbb{D}_{\mathrm{KL}}(q^{\boldsymbol{c}}(\boldsymbol{x}^{w}_{0}|\boldsymbol{x}^{w}_{t},\boldsymbol{x}^{w}_{0}(t))\lVert p^{\boldsymbol{c}}_{\mathrm{ref}}(\boldsymbol{x}^{w}_{0}|\boldsymbol{x}^{w}_{t}))\\
    &-\mathbb{D}_{\mathrm{KL}}(q^{\boldsymbol{c}}(\boldsymbol{x}^{l}_{0}|\boldsymbol{x}^{l}_{t},\boldsymbol{x}^{l}_{0}(t))\lVert p^{\boldsymbol{c}}_{\theta}(\boldsymbol{x}^{l}_{0}|\boldsymbol{x}^{l}_{t}))\\
    &+\mathbb{D}_{\mathrm{KL}}(q^{\boldsymbol{c}}(\boldsymbol{x}^{l}_{0}|\boldsymbol{x}^{l}_{t},\boldsymbol{x}^{l}_{0}(t))\lVert p^{\boldsymbol{c}}_{\mathrm{ref}}(\boldsymbol{x}^{l}_{0}|\boldsymbol{x}^{l}_{t})))
\end{split}    
    \label{eq:lossinversion}
\end{equation}
where $q^{\boldsymbol{c}}(\cdot|\cdot)$ is a Gaussian transition kernel. Using \cref{eq:xtsolution}, the above loss can be simplified as follows:
\begin{equation}
\begin{split}
    \mathcal{L}(\theta)=& -\mathbb{E}_{\substack{(\boldsymbol{x}^{w}_{0},\boldsymbol{x}^{l}_{0}, \boldsymbol{c})\sim \mathcal{D},t \sim \mathcal{U}(0,T),\\ \boldsymbol{x}_{t}^{w} \sim p^{\boldsymbol{c}}_{\theta}(\boldsymbol{x}_{t}^{w}|\boldsymbol{x}_{0}^{w}), \boldsymbol{x}_{t}^{l} \sim p^{\boldsymbol{c}}_{\theta}(\boldsymbol{x}_{t}^{l}|\boldsymbol{x}_{0}^{l})}} \log \sigma (-\beta w(t)(\\ 
    &\lVert \tau^{w}_{t}-\epsilon_{\theta}^{t}(\boldsymbol{x}_{t}^{w},\boldsymbol{c}) \rVert_{2}^{2}-\lVert \tau^{w}_{t}-\epsilon_{\mathrm{ref}}^{t}(\boldsymbol{x}_{t}^{w},\boldsymbol{c}) \rVert_{2}^{2}\\
    &-\lVert \tau^{l}_{t}-\epsilon_{\theta}^{t}(\boldsymbol{x}_{t}^{l},\boldsymbol{c}) \rVert_{2}^{2}+\lVert \tau^{l}_{t}-\epsilon_{\mathrm{ref}}^{t}(\boldsymbol{x}_{t}^{l},\boldsymbol{c}) \rVert_{2}^{2}))
\end{split}    
    \label{eq:lossinversionconcrete}
\end{equation}
where $\tau_{t}^{*}=(\boldsymbol{x}_{0}^{*}(t)-\boldsymbol{x}^{*}_{0})/\sigma_{t}+\delta_{t}(\boldsymbol{x}^{*}_{0}(t))$ (refer to \cref{fig:pipeline}) and $w(t)$ is a weight function defined consistent with \cite{wallace2024diffusion}.

\paragraph{Link to Diffusion-DPO}  
\label{subsec:dpo-diffusion}
Diffusion-DPO defines a reward function on the denoising trajectory, which expands the decision space because it requires combining the reward functions across all denoise steps. This problem is well known to be closely related to delayed feedback/sparse rewards in reinforcement learning, where effective feedback is only available after generating the entire trajectory. Within our framework, Diffusion-DPO utilizes a rough approximation of $\delta_{t}$ which uses i.i.d random noise $\delta_{t}^{\mathrm{DPO}}(\boldsymbol{x}_{0}(t)) \sim \mathcal{N}(0,I)$, matching diffusion incorporated in the forward process of noise addition.
\section{Experiments}
\label{sec:experiments}
\begin{table*}[t]
    \centering
    \begin{tabular}{llcccccccc}
        \toprule
         \multirow{2}{*}{Model}& \multirow{2}{*}{Baselines}& \multicolumn{4}{c}{Win-rate (HPDv2)} & \multicolumn{4}{c}{Win-rate (Parti-Prompts)} \\
         & & Aesthetic  & PickScore & HPS & CLIP  & Aesthetic  & PickScore & HPS & CLIP \\
         \midrule
         \multirow{3}{*}{InPO-SDXL} & vs. Base-SDXL & \textbf{56.37} & \textbf{79.25} & \textbf{85.38} & \textbf{53.37} & \textbf{61.70}  & \textbf{72.89} & \textbf{78.31} & \textbf{50.55}   \\
         & vs. SFT-SDXL&  \textbf{66.66}  & \textbf{88.31} & \textbf{89.91} & \textbf{59.00} & \textbf{69.70}  & \textbf{88.54} & \textbf{88.91} & \textbf{56.99} \\
         & vs. DPO-SDXL& \textbf{55.87} & \textbf{59.28} & \textbf{65.81} & 46.91 & \textbf{57.17}  & \textbf{56.92} & \textbf{60.11} & 42.28  \\
         \midrule
         \multirow{4}{*}{InPO-SD1.5} & vs. Base-SD1.5& \textbf{80.19} & \textbf{85.84} & \textbf{90.22} & \textbf{66.44} & \textbf{74.63}  & \textbf{73.16} & \textbf{81.86} & \textbf{61.89}  \\
         & vs. SFT-SD1.5& \textbf{56.63} & \textbf{66.84} & \textbf{58.91} & \textbf{57.87} & \textbf{52.27}  & \textbf{62.50} & \textbf{58.03} & \textbf{57.54}  \\
         & vs. DPO-SD1.5& \textbf{70.50} & \textbf{74.94} & \textbf{84.06} & \textbf{61.09} & \textbf{68.01}  & \textbf{63.54} & \textbf{75.31} & \textbf{57.54} \\
         & vs. KTO-SD1.5& \textbf{59.81} & \textbf{67.41} & \textbf{60.94} & \textbf{60.03} & \textbf{59.01}  & \textbf{62.25} & \textbf{58.76} & \textbf{57.60}  \\
         \bottomrule 
    \end{tabular}
    \caption{Win-rate comparison between DDIM-InPO and baselines. We use prompts from HPDv2 and Parti-Prompts, evaluated across different evaluators. SDXL and  SD1.5 are used as base models. For simplicity, we denote models using a \textquoteleft Method-Base' format, such as InPO-SDXL to represent the SDXL model fine-tuned with the DDIM-InPO method. Win-rates exceeding 50\% are highlighted in \textbf{bold}. }
    \label{tab:winrate}
\end{table*}
\subsection{Setup}
\paragraph{Datasets and Models}  
We conduct all experiments on the Pick-a-Pic v2 \cite{kirstain2023pick} dataset, which is collected from the Pick-a-Pic web application (refer to \cite{kirstain2023pick} for details). Consistent with the setting in \cite{wallace2024diffusion}, approximately 12\% of the data is excluded, where preferences cannot be determined. As a result, we obtain 851,293 data pairs and 58,960 unique prompts. Our method is applied to fine-tune the base models Stable Diffusion 1.5 (SD1.5) \cite{rombach2022high} and Stable Diffusion XL-base-1.0 (SDXL) \cite{podell2023sdxl}.
\paragraph{Basic Hyperparameters}  
To ensure a fair comparison of training efficiency, we use hyperparameters largely consistent with those in \cite{wallace2024diffusion}. All experimental details are provided in \suppref{sec:suppexperimentdetals}. We employ AdamW \cite{Loshchilov2017FixingWD} as the optimizer for SD1.5 and Adafactor \cite{Shazeer2018AdafactorAL} for SDXL. All experiments are conducted on eight NVIDIA H800 GPUs, with each GPU handling a batch size of one pair and gradient accumulation over 128 steps, resulting in an effective batch size of 1024 pairs (half the batch size compared to \cite{wallace2024diffusion}). The learning rate is set to $\frac{2000}{\beta}2.048^{-8}$, with a linear warm-up phase. For SD1.5 training, $\beta$ is set to 2000, while for SDXL training, it is set to 5000.
\paragraph{Evaluation}  
We compare DDIM-InPO with existing baselines using user studies and automatic preference metrics. We use the following four evaluators to evaluate image quality: the LAION aesthetic classifier \cite{Schuhmann2022LAION5BAO} to assess image aesthetic scores, CLIP \cite{Radford2021LearningTV} to measure text-image alignment, and PickScore \cite{kirstain2023pick} and HPS \cite{Wu2023HumanPS1} to predict human preferences for given image-text pairs. In test stage, we generated images using both the baseline and our model on the Parti-Prompts (1632 prompts) and HPDv2 (3200 prompts) test sets. The median and win-rate (how frequently does the evaluator prefer the generations from DDIM-InPO over those from baselines) of the aforementioned four evaluators are used as automatic preference metrics for comparison. Additionally, we conduct a user study (16 participants) to compare DDIM-InPO with the baselines. We design three questions as \cite{wallace2024diffusion}: \textit{(1)Which image is your overall preferred choice? (2)Which image is more visually attractive? (3)Which image better matches the text description? } We also compare the training gpu hours (on H800) with baselines. In this work, we adopt the following baselines: base model (SD1.5 and SDXL), supervised fine-tuning (SFT), Diffusion-DPO, and Diffusion-KTO (for SD1.5). 
\begin{figure}[t]
  \centering
   \includegraphics[width=1\linewidth]{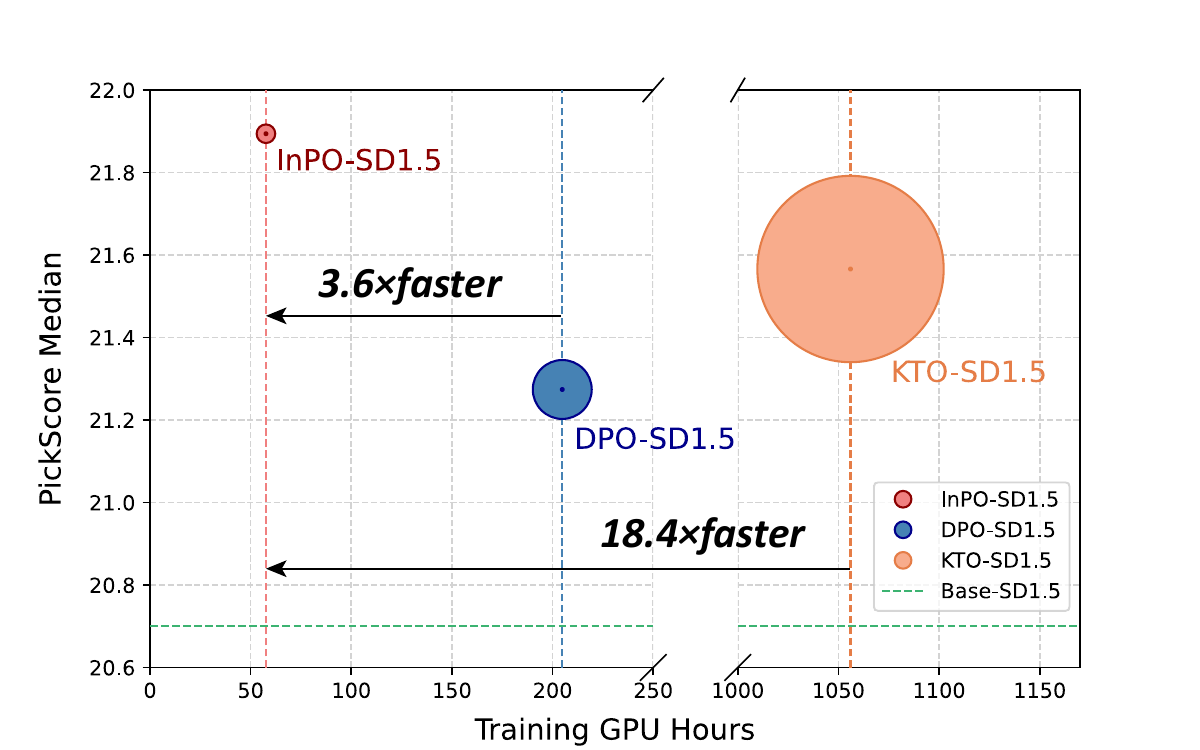}

   \caption{Comparison of the trade-off between the quality of generated images and training efficiency following human preference optimization of SD1.5 on the HPDv2 test set. Sizes of the circles represent the volume of training data used. Our DDIM-InPO achieves superior performance, with a training speed that is 18.4 and 3.6 times faster than Diffusion-KTO \cite{li2024aligning} and Diffusion-DPO \cite{wallace2024diffusion}, respectively, while producing images of higher quality.}
   \label{fig:speedsd15}
\end{figure}

\begin{figure*}[t]
  \centering
   \includegraphics[width=1\linewidth]{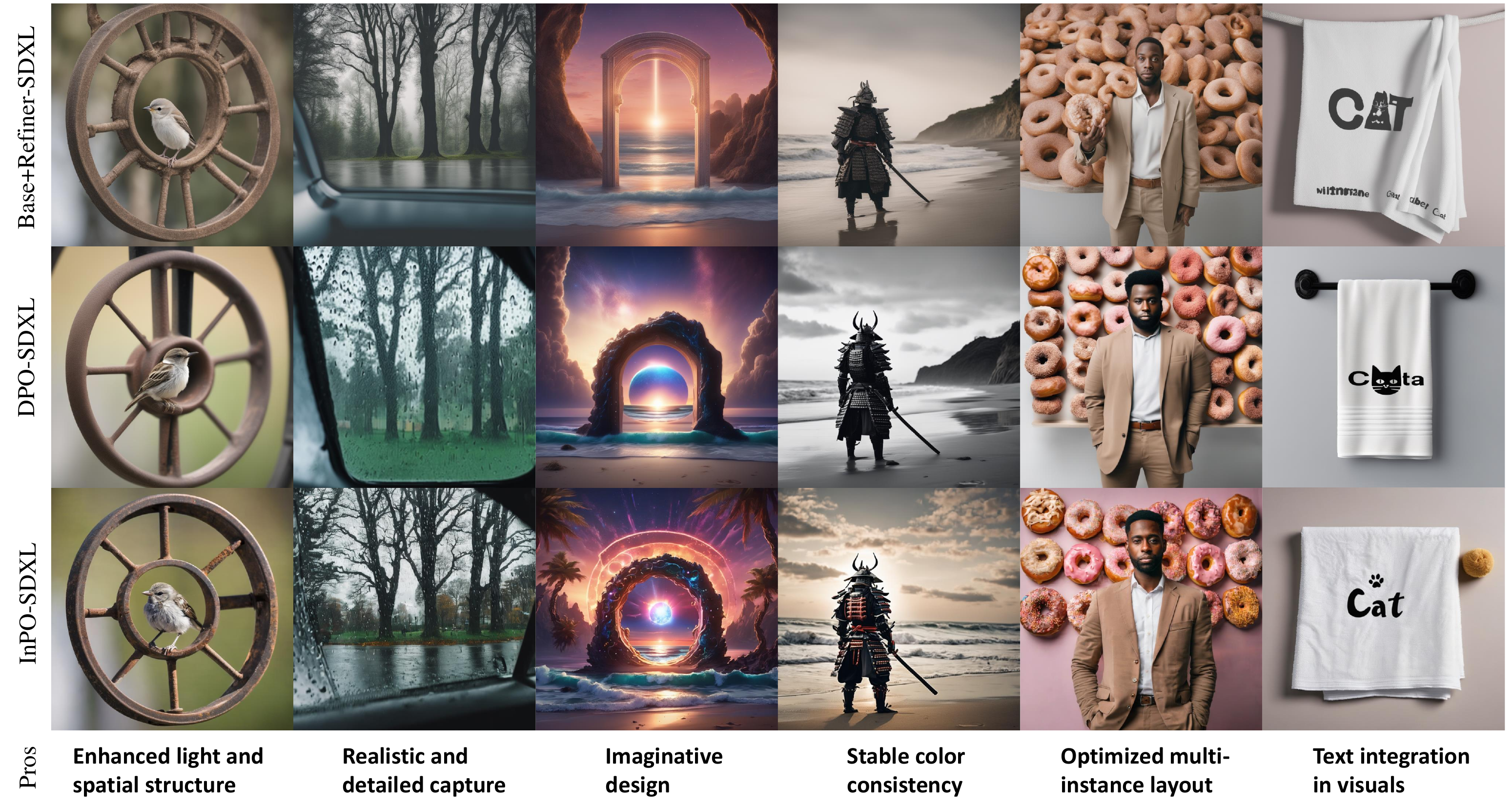}

   \caption{Advantages of DDIM-InPO: Compared with Refinement technique and Diffusion-DPO, we find that images generated by DDIM-InPO exhibit enhanced light and spatial structure, better realism and detail capture, greater imaginative design, stable color consistency, optimized multi-instance layout and text integration in visuals. These are some hidden advantages aligned with human preferences. Prompts from left to right: \textit{(1) A small bird sitting in a metal wheel. (2) Trees seen through a car window on a rainy day. (3) Description, An artistic rendering of a cosmic portal with a beach at dusk on the other side. (4) A person in full samurai armor at the beach. (5) A man standing in front of a bunch of doughnuts. (6) A towel with the word \textquoteleft cat' printed on it, simple and clear text.}}
   \label{fig:inpoadvantage2}
\end{figure*}
\subsection{Primary Results}
\paragraph{Quantitative Results} 
Firstly, we utilize the quantitative evaluation method for model preferences outlined in \cite{li2024aligning, wallace2024diffusion}. \cref{tab:winrate} presents the win-rates of the InPO-aligned diffusion models in comparison to the baselines. Overall, following fine-tuning with DDIM-InPO, both SD1.5 and SDXL achieve superior performance compared to the baselines across nearly all evaluators and test datasets, thereby validating the effectiveness of our method. Specifically, HPS evaluation shows that, InPO-SDXL obtains a win-rate of 65.81\% against DPO-SDXL, while InPO-SD1.5 achieves a win-rate of 60.94\% compared to KTO-SD1.5 on the HPDv2 test set. Moreover, it significantly improves the alignment with human preferences for the base model. According to HPS, it achieves win-rate of up to 85.38\% and 90.22\% when compared to the SD1.5 and SDXL base models, respectively. More quantitative results are available in \suppref{sec:suppquanti}.

It should be noted that our DDIM-InPO significantly advances training efficiency, with a training speed that is 18.4 and 3.6 times faster than Diffusion-KTO and Diffusion-DPO, respectively, as shown in \cref{fig:speedsd15}, while producing higher-quality images. Additionally, \cref{fig:inpoadvantage1} (top) presents the results of our user study, where we find that human evaluators prefer the model aligned with DDIM-InPO. The evaluation results clearly show that our method provides a more efficient way for fine-tuning diffusion models with human preferences, demonstrating the effectiveness of our proposed paradigm for learning specific latent variables. 
\paragraph{Qualitative results} 
We display qualitative comparison results of InPO-SDXL against several baselines (Base-SDXL, DPO-SDXL) in \cref{fig:inpoadvantage1} (bottom). The images generated by InPO-SDXL demonstrate enhanced performance in both textual alignment and visual attractiveness. In addition, \cref{fig:inpoadvantage2} illustrates the advantages of our model (more results in \suppref{sec:suppquanli}), which serves as a method for directly learning human preferences for diffusion models. The generated images showcase improved light and spatial structures, detailed capture, greater creativity, consistent color, organized multi-instance layouts, and text-visual integration. These advantages align closely with the preferences of users.
\begin{figure}[t]
  \centering
   \includegraphics[width=0.9\linewidth]{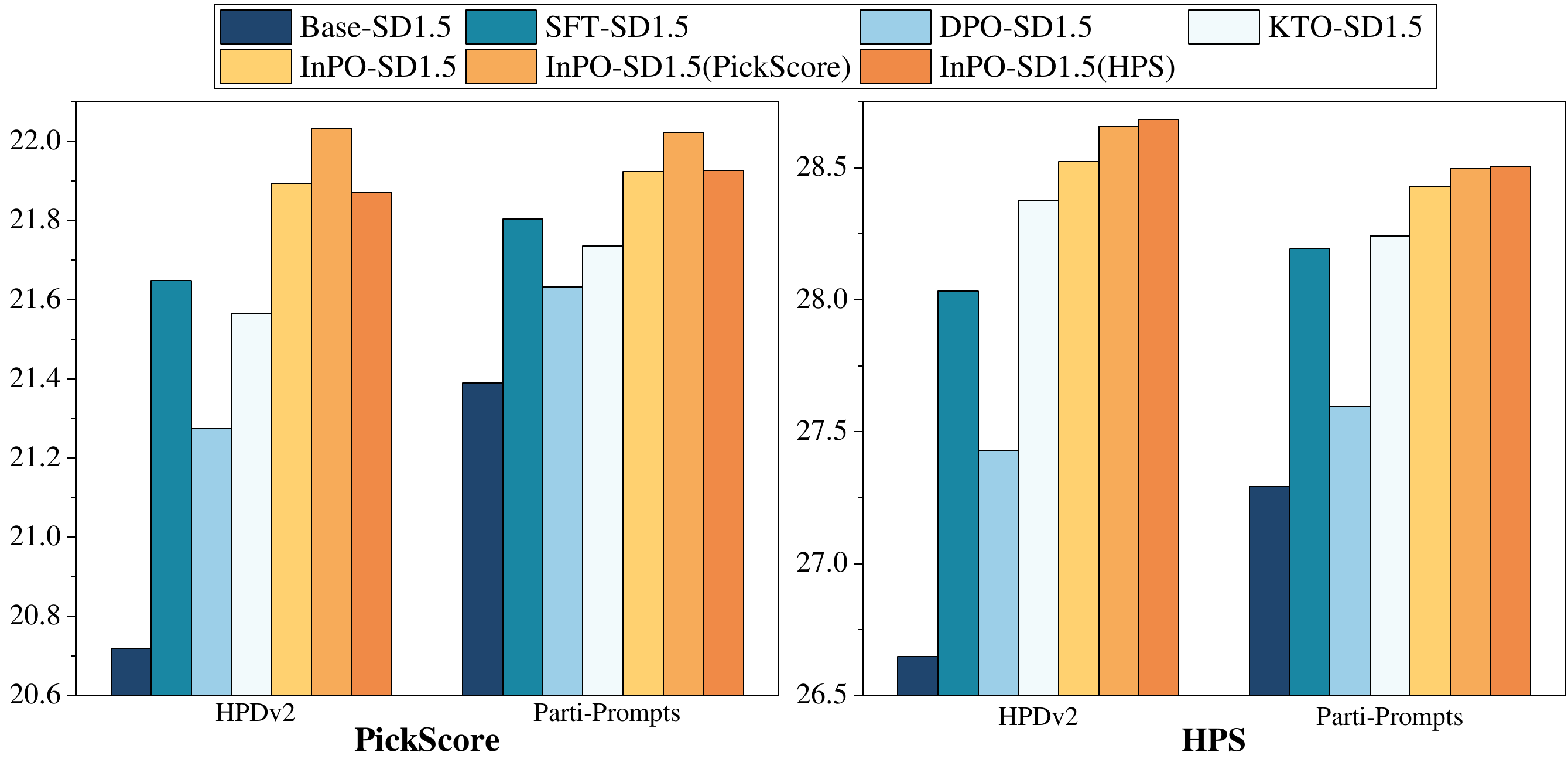}
   \caption{Median of PickScore and HPS comparisons for all baselines and test datasets on SD1.5. (Detailed analysis in \cref{subsec:rlaif})}
   \label{fig:rlaif}
\end{figure}
\subsection{AI Preference Optimization}
\label{subsec:rlaif}
The high cost of manual annotation has driven researchers to seek alternative approaches. Bai \etal \cite{bai2022constitutional} introduce RL from AI feedback as one such solution. It follows naturally that PickScore, and HPS can be employed as AI evaluator to simulate human preferences (further results in \suppref{sec:supprlaif}). We relabel the Pick-a-Pic v2 dataset using AI evaluators and fine-tuned the SD1.5 model. As shown in \cref{fig:rlaif}, we observe a further improvement in these two metrics.
\subsection{Ablations}
\label{subsec:ablation}
\cref{tab:ablation} presents the results of our ablation experiments on SD1.5 and the median values of the preference evaluators on the HPDv2 test set. More ablations are provided in \suppref{sec:suppablation}. For a clearer comparison of our method's effectiveness, we present the results for the Base-SD1.5 and DPO-SD1.5 baselines in rows 3 and 4 of \cref{tab:ablation}.
\begin{table}[t]
\footnotesize
    \centering
    \begin{tabular}{lccccc}
        \toprule
         \multirow{2}{*}{}& \multicolumn{4}{c}{Median (HPDv2)} & \multirow{2}{*}{\begin{tabular}[c]{@{}c@{}}GPU\\ hours\end{tabular}} \\
          \cline{2-5}
         & Aesthetic  & PickScore & HPS & CLIP &\\
         \midrule
         Base-SD1.5 &  5.3491  & 20.719 & 26.647 & 34.276 &-\\
         DPO-SD1.5 &  5.6922  & 21.274 & 27.428 & 35.902 & 204.8\\
         \midrule
         Base initialized &  5.6504  & 21.763 & 28.237 & 36.238 & 57.6\\
         \midrule
         $\beta= 3000$ &  5.7730  & 21.808 & 28.505 & 36.788 & 57.6\\
         $\beta= 4000$ &  5.7637  & 21.796 & 28.448 & 36.685 & 57.6\\
         $\beta= 5000$ &  5.7428  & 21.771 & 28.473 & 36.669 & 57.6\\
         \midrule
         Inv-steps $n$=30 &  \textbf{5.8225}  & \textbf{21.957} & \textbf{28.568} & \textbf{37.019} & 136\\
         Inv-steps $n$=5 &  5.7715  & 21.845 & 28.432 & 36.769 & 43.2 \\
         Inv-steps $n$=3 &  5.7489  & 21.833 & 28.508 & 36.542 & 36\\
         \midrule
         CFG $w_{\mathrm{inv}}$=1 &  5.7381 & 21.835 & 28.516 & 36.451 & 62.4\\
         CFG $w_{\mathrm{inv}}$=5 &  5.7320  & 21.712 & 28.462 & 36.426 & 104\\
         CFG $w_{\mathrm{inv}}$=7.5&  5.7495  & 21.742 & 28.473 & 36.469 & 104\\
         \midrule
         InPO-SD1.5 &   \underline{5.7734}  & \underline{21.894} & \underline{28.523} & \underline{36.876} & 57.6\\
         \bottomrule 
    \end{tabular}
    \caption{Ablation studies of our DDIM-InPO on fine-tuning SD1.5. In each column of preference evaluators, the highest value is highlighted in bold, and the second highest is underlined. For detailed analysis, refer to \cref{subsec:ablation}}
    \label{tab:ablation}
\end{table}
\paragraph{Proposed improvements} 
The fifth row of \cref{tab:ablation} presents the results of initializing $p^{\boldsymbol{c}}_{\theta}$ in \cref{eq:dpoddim} with the SD1.5 base model. After applying our core contribution, DDIM inversion, we observe a significant improvement in quality compared to the median values of the evaluators for the baselines, indicating that enhancement primarily stems from our method. As instructed by \cite{rafailov2024direct}, practically an available SFT model can be used to initialize the reference model $p^{\boldsymbol{c}}_{\mathrm{ref}}$. This strategy aids in reducing the issue of distribution shift.
\paragraph{Parameters Choices} We set $\beta$ to 2000 for SD1.5. Rows 6 to 8 of \cref{tab:ablation} indicate that if $\beta$ is set too high, the KL divergence penalty term severely limits the model's flexibility in adjustments. It also maintains a consistent learning rate with other baselines, ensuring a fair comparison in terms of efficiency. Rows 9 to 11 of \cref{tab:ablation} show the steps required for DDIM inversion. We observe that setting inversion steps to 30 yields the best performance, further confirming the effectiveness of our method. However, this also increases the training time. Thus we select inversion steps $n=10$ as it strikes a good balance between generation quality and training speed. The work of \cite{miyake2023negative, mokady2023null} indicates that DDIM inversion can lead to considerable numerical errors when Classifier Free Guidance (CFG) \cite{Ho2022ClassifierFreeDG} $w_{\mathrm{inv}}>0$. Rows 12 to 14 of \cref{tab:ablation} demonstrate that a CFG $w_{\mathrm{inv}}>0$ during the inversion process adversely impacts the fine-tuning of human preferences. Consequently, we utilize unconditional DDIM inversion, setting CFG $w_{\mathrm{inv}}=0$. We optimize parameters to balance generation quality and training efficiency.

\subsection{Conditional generation}
\begin{figure}[t]
  \centering
   \includegraphics[width=1\linewidth]{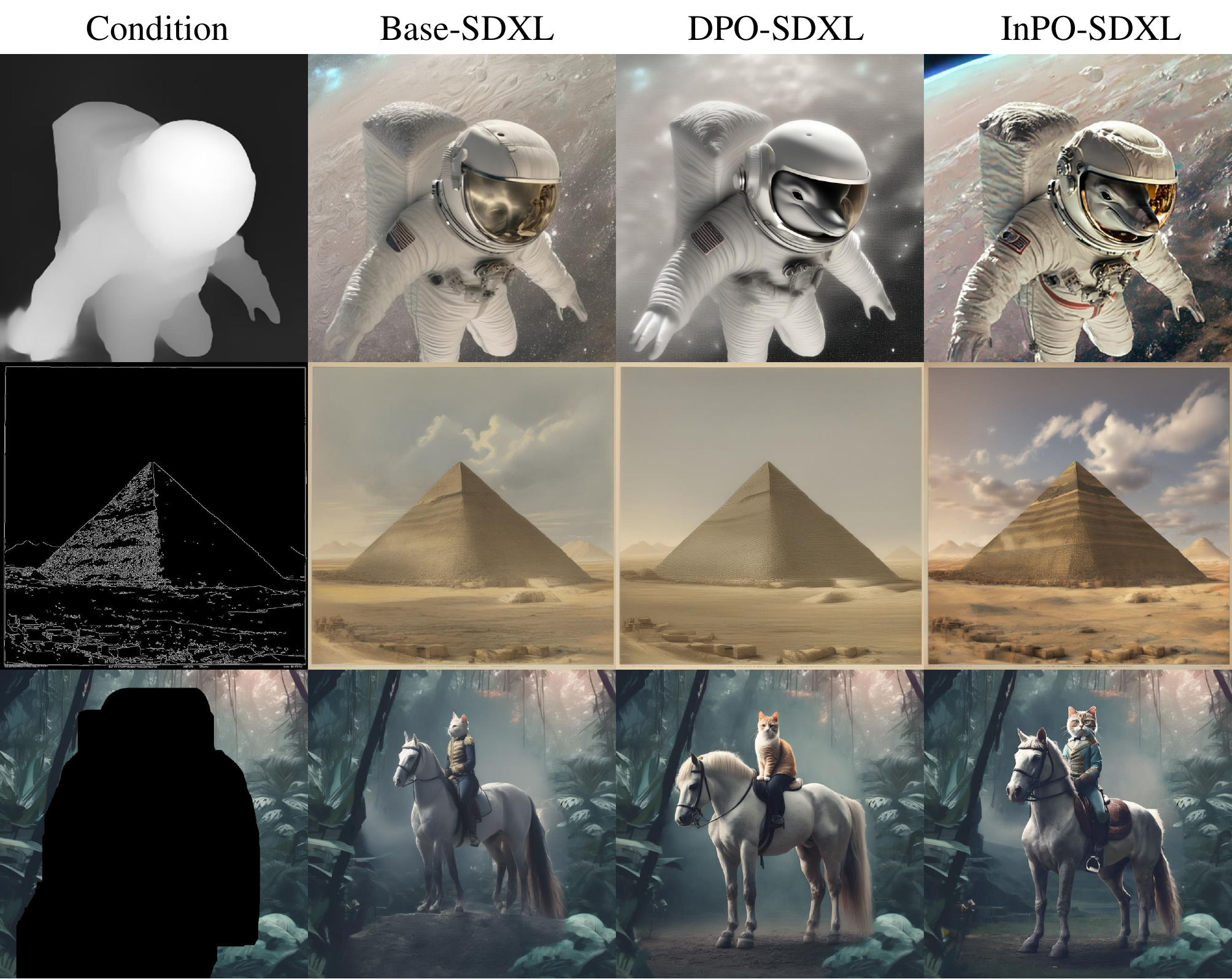}

   \caption{Qualitative evaluation of InPO-SDXL in comparison with Base-SDXL and DPO-SDXL on conditional generation tasks (From top to bottom: depth map, canny edge, and inpainting). Prompts: \textit{(1) A dolphin in an astronaut suit on Saturn. (2) The Great Pyramid of Giza. (3) A cat riding on a horse.}}
   \label{fig:controlnet}
\end{figure}

After fine-tuning the model using the DDIM-InPO method, it can be directly applied to conditional generation tasks without the need for further training. We add conditional control to InPO-SDXL using ControlNet \cite{Zhang2023AddingCC} which originally trained with Base-SDXL. Specifically, we use depth map and canny edge as conditions for text-to-image generation tasks. Furthermore, we evaluate our model on image inpainting \cite{Meng2021SDEditGI} tasks. As shown in \cref{fig:controlnet}, the advantages of our model are seamlessly transferred to conditional generation tasks, demonstrating superior generation quality compared with Base-SDXL and DPO-SDXL.

\subsection{Discussion}
Since we conduct experiments on a given dataset Pick-a-Pic v2, the data is not generated by the diffusion model itself. The work of \cite{wallace2024diffusion} indicates that the quality of the Pick-a-Pic v2 dataset falls between SDXL and SD1.5, as it is sourced from SDXL-Beta and Dreamlike. Our experimental results show that supervised fine-tuning SD1.5 enhances its performance, whereas any level of fine-tuning on SDXL results in a decline in model metrics. Therefore, we recommend maxmizing the likelihood of preferred pairs $(\boldsymbol{x}^{w},\boldsymbol{c})$ before aligning the SD1.5 with human preferences, that is, $p^{\boldsymbol{c}}_{\mathrm{ref}}=\arg \max_{p}\mathbb{E}_{(\boldsymbol{x}^{w},\boldsymbol{c})\sim \mathcal{D}}[\log p(\boldsymbol{x}^{w}|\boldsymbol{c})]$. In contrast, the SDXL model utilizes the parameters of itself.

\section{Conclusion and Limitations}
\label{sec:conclusion}
We introduce DDIM-InPO, an efficient method for aligning diffusion models with human preferences, focusing on directly fine-tuning specific latent variables that are strongly correlated with the target image. By fine-tuning the T2I diffusion models using the Pick-a-Pic v2 dataset, our experiments show that DDIM-InPO significantly enhances training efficiency, achieving state-of-the-art performance in just 400 training steps and surpassing all baseline models across nearly all evaluators. 

\paragraph{Limitations} DDIM-InPO is based on offline fine-tuning and is sensitive to dataset quality; the presence of harmful, violent, and pornographic content can affect the results, leading to the generation of inappropriate outputs, such as overly feminine images. In addition, we consider the online training paradigm to be a promising and reliable approach for the future. \suppref{sec:suppdiscussion} includes further discussion.
\section*{Acknowledgments}
\label{sec:acknowledgments}
This work was supported by the National Major Science and Technology Projects (the grant number 2022ZD0117000)  and the National Natural Science Foundation of China (grant number 62202426).

{
    \small
    \bibliographystyle{ieeenat_fullname}

}

\newpage
\clearpage
\setcounter{page}{1}
\setcounter{section}{0}


\onecolumn  
\begin{center}
    \textbf{\Large InPO: Inversion Preference Optimization with Reparameterized DDIM for} \\[1ex]
    \textbf{\Large Efficient Diffusion Model Alignment} \\[2ex]
    \Large Supplementary Material
\end{center}

\suppsection{Background}
\label{sec:supprelatedworks}
\paragraph{Conditional Generative Models}
Inspired by non-equilibrium thermodynamics, diffusion models gradually introduce noise into the data and learn to reverse this process starting from pure noise, ultimately generating target data that aligns with the original data distribution. Broadly, diffusion models can be categorized into two types: denoising diffusion and score-matching \cite{Song2020ScoreBasedGM}. Diffusion models \cite{Karras2022ElucidatingTD, Dhariwal2021DiffusionMB, Peebles2022ScalableDM}  have become the leading approach in generative modeling \cite{Balaji2022eDiffITD, Wang2023CogVLMVE, Zhai2021ScalingVT}, outperforming earlier methods like Generative Adversarial Networks (GANs) \cite{Reed2016GenerativeAT} and Variational Autoencoders (VAEs) in both quality and stability. They have shown outstanding results across a wide range of generative tasks, including image \cite{BetkerImprovingIG, Sauer2023AdversarialDD, Chen2023PixArtFT, Pernias2023WuerstchenAE} and video generation \cite{Singer2022MakeAVideoTG, Esser2023StructureAC, Gupta2023PhotorealisticVG, Ho2022ImagenVH, Esser2023StructureAC, Blattmann2023AlignYL, Blattmann2023StableVD}. This paper primarily focuses on diffusion models for conditional image generation \cite{Dai2023EmuEI}, encompassing tasks such as text-to-image synthesis, additional control conditions, and image inpainting for restoration. Conditional image generation\cite{ Meng2021SDEditGI, miyake2023negative, Huang2024DiffusionMI, Hertz2022PrompttoPromptIE} leverages guidance conditions to synthesize new images from scratch, with the conditions being either single or multiple. Earlier approaches predominantly relied on class-conditional generation, which required training extra classifiers and utilizing classifier-induced gradients for image synthesis. In contrast, Ho et al. introduce classifier-free guidance \cite{Ho2022ClassifierFreeDG}, which eliminates the need for classifier training and allows for more flexible conditioning. This approach also enables control over the degree of guidance, such as text prompts, by adjusting specific coefficients. Beyond text-based prompts \cite{Saharia2022PhotorealisticTD, Balaji2022eDiffITD}, more specific conditions \cite{Garibi2024ReNoiseRI, Ghosh2023GenEvalAO} can be utilized to achieve finer control over image synthesis. For instance, ControlNet \cite{Zhang2023AddingCC} allows the integration of additional input types, such as depth maps, precise edges, poses, and sketches, to guide the generation process more accurately. Moreover, image restoration is a vital task in computer vision that focuses on enhancing the quality of images affected by various degradations. An example is inpainting \cite{Sheynin2023EmuEP}, where the goal is to fill in missing regions of an image to restore its completeness.

\paragraph{Large Language Models Alignment}
Reinforcement Learning from Human Feedback (RLHF) \cite{Christiano2017DeepRL} is a widely used approach to align models with human preferences. It involves first training a reward model on data that explicitly reflects human preferences, followed by using reinforcement learning techniques to optimize the policy/model, aiming to maximize the reward. As is well recognized, the widely popular model ChatGPT leverages RLHF techniques. A pioneering study, the work of \cite{Ouyang2022TrainingLM}, is the first to apply RLHF for fine-tuning LLMs, which has since gained substantial recognition. Proximal Policy Optimization (PPO) \cite{Schulman2017ProximalPO} is a crucial algorithm in RL, but its training process often requires the simultaneous use of a training model, a reference model, a reward model, and a critic, which is particularly demanding in terms of memory consumption, especially when applied to LLMs \cite{Ouyang2022TrainingLM}. Recent research suggests that it is possible to circumvent traditional RL algorithms. For instance, RAFT \cite{Dong2023RAFTRR} achieves optimization by fine-tuning on online samples with the highest rewards. Meanwhile, RRHF \cite{Yuan2023RRHFRR} aligns models using ranking loss, learning from responses sampled from multiple sources to enhance alignment. The work of \cite{Liu2023StatisticalRS} introduce rejection sampling optimization, where preference data is collected using a reward model to guide the sampling process. DPO bypasses the need for training an explicit reward model by directly optimizing the optimal policy, assuming that pairwise preferences can be approximated using pointwise rewards. To address potential overfitting to preference datasets in DPO, the work of \cite{Azar2023AGT} propose Identity Preference Optimization. The work of \cite{Hong2024ORPOMP} introduce odds ratio preference optimization (ORPO), which incorporates SFT on preference data. In contrast, the work of \cite{Ethayarajh2024KTOMA} avoids reliance on pairwise preference data by combining Kahneman-Tversky optimization (KTO), focusing on directly optimizing utility instead of maximizing the log-likelihood of preferences. Additionally, the work of \cite{Song2023PreferenceRO} propose preference ranking optimization (PRO), which leverages higher-order information embedded in list rankings. However, due to the unique characteristics of diffusion models, these methods cannot be directly applied without significant adaptation.

\paragraph{Additional Diffusion Models Alignment}
Aligning diffusion models with human preferences has recently attracted significant attention. The work of \cite{Prabhudesai2024VideoDA} extend this concept to video diffusion models but encountered challenges, such as the linear increase in reward feedback costs due to the added time dimension. To overcome these issues, they optimize the video diffusion model using gradients obtained from publicly available pre-trained reward models. In contrast, InstructVideo \cite{Yuan2023InstructVideoIV} fine-tunes text-to-video diffusion models based on human feedback rewards, employs partial DDIM sampling to reduce computational costs, and leverages image reward models to enhance video quality while maintaining the model’s generalization ability. Moreover, reward models \cite{Xu2023ImageRewardLA} and timestep-aware alignment methods \cite{Liang2024StepawarePO} for diffusion models warrant further investigation. Human preference alignment techniques developed for LLMs can potentially be adapted for diffusion models. We believe that advances in methods such as IPO, ORPO, and PRO could be extended to diffusion models, potentially enhancing their performance. However, due to the fundamental differences in architecture between LLMs and diffusion models, directly applying LLM techniques to diffusion models may not produce the same level of benefits.

\suppsection{Details of the Primary Derivation} 
\label{sec:suppprimaryderivation}
In this section, we present a detailed derivation of the proposed method. From \cref{eq:rlhf}, we can derive the following:
\begin{equation}
\begin{aligned}
    & \max_{p_{\theta}} \mathbb{E}_{\boldsymbol{x}_{0}\sim p_{\theta}(\boldsymbol{x}_{0}|\boldsymbol{c})}[r(\boldsymbol{x}_{0},\boldsymbol{c})]/\beta- \mathbb{D}_{\mathrm{KL}}[p_{\theta}(\boldsymbol{x}_{0}|\boldsymbol{c})||p_{\mathrm{ref}}(\boldsymbol{x}_{0}|\boldsymbol{c})] \\
    = & \min_{p_{\theta}^{\boldsymbol{c}}} -\mathbb{E}_{\boldsymbol{x}_{0}\sim p_{\theta}^{\boldsymbol{c}}(\boldsymbol{x}_{0})}[r(\boldsymbol{x}_{0},\boldsymbol{c})]/\beta+ \mathbb{D}_{\mathrm{KL}}[p_{\theta}^{\boldsymbol{c}}(\boldsymbol{x}_{0})||p^{\boldsymbol{c}}_{\mathrm{ref}}(\boldsymbol{x}_{0})]\\
    \leq & \min_{p_{\theta}^{\boldsymbol{c}}} -\mathbb{E}_{\boldsymbol{x}_{0}\sim p_{\theta}^{\boldsymbol{c}}(\boldsymbol{x}_{0})}[r(\boldsymbol{x}_{0},\boldsymbol{c})]/\beta+ \mathbb{D}_{\mathrm{KL}}[p_{\theta}^{\boldsymbol{c}}(\boldsymbol{x}_{0},\boldsymbol{x}_{t})||p^{\boldsymbol{c}}_{\mathrm{ref}}(\boldsymbol{x}_{0},\boldsymbol{x}_{t})] \\
    = & \min_{p_{\theta}^{\boldsymbol{c}}} -\mathbb{E}_{p^{\boldsymbol{c}}_{\theta}(\boldsymbol{x}_{0},\boldsymbol{x}_{t})}[r^{\boldsymbol{c}}_{t}(\boldsymbol{x}_{0},\boldsymbol{x}_{t})]/\beta+ \mathbb{D}_{\mathrm{KL}}[p_{\theta}^{\boldsymbol{c}}(\boldsymbol{x}_{0},\boldsymbol{x}_{t})||p^{\boldsymbol{c}}_{\mathrm{ref}}(\boldsymbol{x}_{0},\boldsymbol{x}_{t})] \\
    = & \min_{p_{\theta}^{\boldsymbol{c}}} \mathbb{E}_{p^{\boldsymbol{c}}_{\theta}(\boldsymbol{x}_{0},\boldsymbol{x}_{t})}\left(\log \frac{p^{\boldsymbol{c}}_{\theta}(\boldsymbol{x}_{0},\boldsymbol{x}_{t})}{p^{\boldsymbol{c}}_{\mathrm{ref}}(\boldsymbol{x}_{0},\boldsymbol{x}_{t})\exp (r^{\boldsymbol{c}}_{t}(\boldsymbol{x}_{0},\boldsymbol{x}_{t})/\beta)/Z_{t}(\boldsymbol{c})}-\log Z_{t}(\boldsymbol{c})  \right) \\
    = & \min_{p_{\theta}^{\boldsymbol{c}}} \mathbb{D}_{\mathrm{KL}}(p^{\boldsymbol{c}}_{\theta}(\boldsymbol{x}_{0},\boldsymbol{x}_{t})||p^{\boldsymbol{c}}_{\mathrm{ref}}(\boldsymbol{x}_{0},\boldsymbol{x}_{t})\exp (r^{\boldsymbol{c}}_{t}(\boldsymbol{x}_{0},\boldsymbol{x}_{t})/\beta)/Z_{t}(\boldsymbol{c}))
\end{aligned}
    \label{eq:suppderivation1}
\end{equation}
where $Z_{t}(\boldsymbol{c}) = \sum_{\boldsymbol{x}_{0}}p^{\boldsymbol{c}}_{\mathrm{ref}}(\boldsymbol{x}_{0},\boldsymbol{x}_{t})\exp (r(\boldsymbol{x}_{0},\boldsymbol{c})/\beta)$ is the timestep-aware partition function. From the preceding equation \cref{eq:suppderivation1}, we can express the closed-form solution for the optimal policy $p^{\boldsymbol{c}}_{\theta^{*}}(\boldsymbol{x}_{0},\boldsymbol{x}_{t})$ at timestep $t$:
\begin{equation}
    p^{\boldsymbol{c}}_{\theta^{*}}(\boldsymbol{x}_{0},\boldsymbol{x}_{t}) = p^{\boldsymbol{c}}_{\mathrm{ref}}(\boldsymbol{x}_{0},\boldsymbol{x}_{t})\exp (r^{\boldsymbol{c}}_{t}(\boldsymbol{x}_{0},\boldsymbol{x}_{t})/\beta)/Z_{t}(\boldsymbol{c})
    \label{eq:suppoptimalsolution}
\end{equation}
A straightforward transformation of \cref{eq:suppoptimalsolution} leads to the solution for the \textquoteleft joint' reward at timestep $t$:
\begin{equation}
    r^{\boldsymbol{c}}_{t}(\boldsymbol{x}_{0},\boldsymbol{x}_{t}) = \beta \log \frac{p^{\boldsymbol{c}}_{\theta^{*}}(\boldsymbol{x}_{0},\boldsymbol{x}_{t})}{p^{\boldsymbol{c}}_{\mathrm{ref}}(\boldsymbol{x}_{0},\boldsymbol{x}_{t})}+\beta \log Z_{t}(\boldsymbol{c})
    \label{eq:suppjointreward}
\end{equation}
Then from \cref{eq:rewardfunction}, we can derive the expression for the \textquoteleft initial' reward:
\begin{equation}
    r(\boldsymbol{x}_{0},\boldsymbol{c}) = \beta \mathbb{E}_{p^{\boldsymbol{c}}_{\theta}(\boldsymbol{x}_{t}|\boldsymbol{x}_{0})}\left[\log \frac{p^{\boldsymbol{c}}_{\theta^{*}}(\boldsymbol{x}_{0},\boldsymbol{x}_{t})}{p^{\boldsymbol{c}}_{\mathrm{ref}}(\boldsymbol{x}_{0},\boldsymbol{x}_{t})}\right] + \beta \log Z_{t}(\boldsymbol{c})
    \label{eq:supprewardfunction}
\end{equation}
By reparameterizing this reward and substituting it into the maximum likelihood objective of the Bradley-Terry model, as described in \cref{eq:btmodel}, we derive a maximum likelihood objective defined for a timestep-aware single-step diffusion model. At timestep $t$ it is expressed as:
\begin{equation}
    \mathcal{L}_{t}(\theta):= -\log \sigma \left(\beta\mathbb{E}_{\boldsymbol{x}_{t}^{w} \sim p^{\boldsymbol{c}}_{\theta}(\boldsymbol{x}_{t}^{w}|\boldsymbol{x}_{0}^{w}),\boldsymbol{x}_{t}^{l} \sim p^{\boldsymbol{c}}_{\theta}(\boldsymbol{x}_{t}^{l}|\boldsymbol{x}_{0}^{l})}\left[ \log \frac{p^{\boldsymbol{c}}_{\theta}(\boldsymbol{x}_{0}^{w},\boldsymbol{x}_{t}^{w})}{p^{\boldsymbol{c}}_{\mathrm{ref}}(\boldsymbol{x}_{0}^{w},\boldsymbol{x}_{t}^{w})}-\log \frac{p^{\boldsymbol{c}}_{\theta}(\boldsymbol{x}_{0}^{l},\boldsymbol{x}_{t}^{l})}{p^{\boldsymbol{c}}_{\mathrm{ref}}(\boldsymbol{x}_{0}^{l},\boldsymbol{x}_{t}^{l})}\right]\right)  
    \label{eq:suppdpoddim1}
\end{equation}
To ensure training stability, we accounted for all time steps, leading to the following result:
\begin{equation}
\mathcal{L}(\theta):= -\mathbb{E}_{t,(\boldsymbol{x}^{w}_{0},\boldsymbol{x}^{l}_{0},\boldsymbol{c})\sim \mathcal{D}}\log \sigma 
\left(\beta\mathbb{E}_{\boldsymbol{x}_{t}^{w} \sim p^{\boldsymbol{c}}_{\theta}(\boldsymbol{x}_{t}^{w}|\boldsymbol{x}_{0}^{w}), \boldsymbol{x}_{t}^{l} \sim p^{\boldsymbol{c}}_{\theta}(\boldsymbol{x}_{t}^{l}|\boldsymbol{x}_{0}^{l})}\left[ \log \frac{p^{\boldsymbol{c}}_{\theta}(\boldsymbol{x}_{0}^{w},\boldsymbol{x}_{t}^{w})}{p^{\boldsymbol{c}}_{\mathrm{ref}}(\boldsymbol{x}_{0}^{w},\boldsymbol{x}_{t}^{w})}-\log \frac{p^{\boldsymbol{c}}_{\theta}(\boldsymbol{x}_{0}^{l},\boldsymbol{x}_{t}^{l})}{p^{\boldsymbol{c}}_{\mathrm{ref}}(\boldsymbol{x}_{0}^{l},\boldsymbol{x}_{t}^{l})}\right]\right)    
    \label{eq:suppdpoddim2}
\end{equation}
where $\boldsymbol{x}_{0}^{w}$, $\boldsymbol{x}_{0}^{l}$ are from preference dataset.

\paragraph{Preference Optimization via Inversion}
For the sake of simplicity, we approximate $p^{\boldsymbol{c}}_{\mathrm{ref}}(\boldsymbol{x}_{0},\boldsymbol{x}_{t})$ with $p^{\boldsymbol{c}}_{\mathrm{ref}}(\boldsymbol{x}_{0}|\boldsymbol{x}_{t})p^{\boldsymbol{c}}_{\theta}(\boldsymbol{x}_{t})$, Consequently, the above equation can be simplified as:
\begin{equation}
\begin{aligned}
\mathcal{L}(\theta): &= -\mathbb{E}_{t,(\boldsymbol{x}^{w}_{0},\boldsymbol{x}^{l}_{0},\boldsymbol{c})\sim \mathcal{D}}\log \sigma 
\left(\beta\mathbb{E}_{\boldsymbol{x}_{t}^{w} \sim p^{\boldsymbol{c}}_{\theta}(\boldsymbol{x}_{t}^{w}|\boldsymbol{x}_{0}^{w}), \boldsymbol{x}_{t}^{l} \sim p^{\boldsymbol{c}}_{\theta}(\boldsymbol{x}_{t}^{l}|\boldsymbol{x}_{0}^{l})}\left[ \log \frac{p^{\boldsymbol{c}}_{\theta}(\boldsymbol{x}_{0}^{w},\boldsymbol{x}_{t}^{w})}{p^{\boldsymbol{c}}_{\mathrm{ref}}(\boldsymbol{x}_{0}^{w},\boldsymbol{x}_{t}^{w})}-\log \frac{p^{\boldsymbol{c}}_{\theta}(\boldsymbol{x}_{0}^{l},\boldsymbol{x}_{t}^{l})}{p^{\boldsymbol{c}}_{\mathrm{ref}}(\boldsymbol{x}_{0}^{l},\boldsymbol{x}_{t}^{l})}\right]\right) \\
&= -\mathbb{E}_{t,(\boldsymbol{x}^{w}_{0},\boldsymbol{x}^{l}_{0},\boldsymbol{c})\sim \mathcal{D}}\log \sigma 
\left(\beta\mathbb{E}_{\boldsymbol{x}_{t}^{w} \sim p^{\boldsymbol{c}}_{\theta}(\boldsymbol{x}_{t}^{w}|\boldsymbol{x}_{0}^{w}), \boldsymbol{x}_{t}^{l} \sim p^{\boldsymbol{c}}_{\theta}(\boldsymbol{x}_{t}^{l}|\boldsymbol{x}_{0}^{l})}\left[ \log \frac{p^{\boldsymbol{c}}_{\theta}(\boldsymbol{x}^{w}_{0}|\boldsymbol{x}^{w}_{t})}{p^{\boldsymbol{c}}_{\mathrm{ref}}(\boldsymbol{x}^{w}_{0}|\boldsymbol{x}^{w}_{t})}-\log \frac{p^{\boldsymbol{c}}_{\theta}(\boldsymbol{x}^{l}_{0}|\boldsymbol{x}^{l}_{t})}{p^{\boldsymbol{c}}_{\mathrm{ref}}(\boldsymbol{x}^{l}_{0}|\boldsymbol{x}^{l}_{t})}\right]\right)
\end{aligned}
    \label{eq:suppdpoddim3}
\end{equation}
Here, we introduce a Gaussian probability density transition function $q^{\boldsymbol{c}}(\cdot |\cdot)$. It is evident that $q^{\boldsymbol{c}}(\boldsymbol{x}_{0}|\boldsymbol{x}_{t},\boldsymbol{x}_{0})=1$, and we have:
\begin{equation}
\begin{aligned}
\mathcal{L}(\theta): = -\mathbb{E}_{t,(\boldsymbol{x}^{w}_{0},\boldsymbol{x}^{l}_{0},\boldsymbol{c})\sim \mathcal{D}} & \log \sigma 
(\beta\mathbb{E}_{\boldsymbol{x}_{t}^{w} \sim p^{\boldsymbol{c}}_{\theta}(\boldsymbol{x}_{t}^{w}|\boldsymbol{x}_{0}^{w}), \boldsymbol{x}_{t}^{l} \sim p^{\boldsymbol{c}}_{\theta}(\boldsymbol{x}_{t}^{l}|\boldsymbol{x}_{0}^{l})} \\ &\mathbb{E}_{\boldsymbol{x}^{w}_{0} \sim q^{\boldsymbol{c}}(\boldsymbol{x}^{w}_{0}|\boldsymbol{x}^{w}_{t},\boldsymbol{x}^{w}_{0}), \boldsymbol{x}^{l}_{0} \sim q^{\boldsymbol{c}}(\boldsymbol{x}^{l}_{0}|\boldsymbol{x}^{l}_{t},\boldsymbol{x}^{l}_{0})}\left[ \log \frac{p^{\boldsymbol{c}}_{\theta}(\boldsymbol{x}^{w}_{0}|\boldsymbol{x}^{w}_{t})}{p^{\boldsymbol{c}}_{\mathrm{ref}}(\boldsymbol{x}^{w}_{0}|\boldsymbol{x}^{w}_{t})}-\log \frac{p^{\boldsymbol{c}}_{\theta}(\boldsymbol{x}^{l}_{0}|\boldsymbol{x}^{l}_{t})}{p^{\boldsymbol{c}}_{\mathrm{ref}}(\boldsymbol{x}^{l}_{0}|\boldsymbol{x}^{l}_{t})}\right])
\end{aligned}
    \label{eq:suppdpoddim4}
\end{equation}
In this context, $\boldsymbol{x}_{0}(t)$ serves as a reliable approximation of $\boldsymbol{x}_{0}$ at timestep $t$, that is $q^{\boldsymbol{c}}(\boldsymbol{x}^{w}_{0}|\boldsymbol{x}^{w}_{t},\boldsymbol{x}^{w}_{0}) \approx q^{\boldsymbol{c}}(\boldsymbol{x}^{w}_{0}|\boldsymbol{x}^{w}_{t},\boldsymbol{x}^{w}_{0}(t))$. Consequently, \cref{eq:suppdpoddim4} can be estimated as:
\begin{equation}
\begin{aligned}
\mathcal{L}(\theta): \approx -\mathbb{E}_{t,(\boldsymbol{x}^{w}_{0},\boldsymbol{x}^{l}_{0},\boldsymbol{c})\sim \mathcal{D}} & \log \sigma 
(\beta\mathbb{E}_{\boldsymbol{x}_{t}^{w} \sim p^{\boldsymbol{c}}_{\theta}(\boldsymbol{x}_{t}^{w}|\boldsymbol{x}_{0}^{w}), \boldsymbol{x}_{t}^{l} \sim p^{\boldsymbol{c}}_{\theta}(\boldsymbol{x}_{t}^{l}|\boldsymbol{x}_{0}^{l})} \\ &\mathbb{E}_{\boldsymbol{x}^{w}_{0} \sim q^{\boldsymbol{c}}(\boldsymbol{x}^{w}_{0}|\boldsymbol{x}^{w}_{t},\boldsymbol{x}^{w}_{0}(t)), \boldsymbol{x}^{l}_{0} \sim q^{\boldsymbol{c}}(\boldsymbol{x}^{l}_{0}|\boldsymbol{x}^{l}_{t},\boldsymbol{x}^{l}_{0}(t))}\left[ \log \frac{p^{\boldsymbol{c}}_{\theta}(\boldsymbol{x}^{w}_{0}|\boldsymbol{x}^{w}_{t})}{p^{\boldsymbol{c}}_{\mathrm{ref}}(\boldsymbol{x}^{w}_{0}|\boldsymbol{x}^{w}_{t})}-\log \frac{p^{\boldsymbol{c}}_{\theta}(\boldsymbol{x}^{l}_{0}|\boldsymbol{x}^{l}_{t})}{p^{\boldsymbol{c}}_{\mathrm{ref}}(\boldsymbol{x}^{l}_{0}|\boldsymbol{x}^{l}_{t})}\right])
\end{aligned}
    \label{eq:suppdpoddim5}
\end{equation}
Then according to Jensen’s inequality, we can obtain:
\begin{equation}
\begin{aligned}
\mathcal{L}(\theta): \leq &-\mathbb{E}_{t,(\boldsymbol{x}^{w}_{0},\boldsymbol{x}^{l}_{0},\boldsymbol{c})\sim \mathcal{D}} \mathbb{E}_{\boldsymbol{x}_{t}^{w} \sim p^{\boldsymbol{c}}_{\theta}(\boldsymbol{x}_{t}^{w}|\boldsymbol{x}_{0}^{w}), \boldsymbol{x}_{t}^{l} \sim p^{\boldsymbol{c}}_{\theta}(\boldsymbol{x}_{t}^{l}|\boldsymbol{x}_{0}^{l})}\log \sigma 
(\beta\\ &\mathbb{E}_{\boldsymbol{x}^{w}_{0} \sim q^{\boldsymbol{c}}(\boldsymbol{x}^{w}_{0}|\boldsymbol{x}^{w}_{t},\boldsymbol{x}^{w}_{0}(t)), \boldsymbol{x}^{l}_{0} \sim q^{\boldsymbol{c}}(\boldsymbol{x}^{l}_{0}|\boldsymbol{x}^{l}_{t},\boldsymbol{x}^{l}_{0}(t))}\left[ \log \frac{p^{\boldsymbol{c}}_{\theta}(\boldsymbol{x}^{w}_{0}|\boldsymbol{x}^{w}_{t})}{p^{\boldsymbol{c}}_{\mathrm{ref}}(\boldsymbol{x}^{w}_{0}|\boldsymbol{x}^{w}_{t})}-\log \frac{p^{\boldsymbol{c}}_{\theta}(\boldsymbol{x}^{l}_{0}|\boldsymbol{x}^{l}_{t})}{p^{\boldsymbol{c}}_{\mathrm{ref}}(\boldsymbol{x}^{l}_{0}|\boldsymbol{x}^{l}_{t})}\right])\\
&=-\mathbb{E}_{\substack(\boldsymbol{x}^{w}_{0},\boldsymbol{x}^{l}_{0}, \boldsymbol{c})\sim \mathcal{D},t \sim \mathcal{U}(0,T),  \boldsymbol{x}_{t}^{w}  \sim p^{\boldsymbol{c}}_{\theta}(\boldsymbol{x}_{t}^{w}|\boldsymbol{x}_{0}^{w}), \boldsymbol{x}_{t}^{l} \sim p^{\boldsymbol{c}}_{\theta}(\boldsymbol{x}_{t}^{l}|\boldsymbol{x}_{0}^{l})} \log \sigma (-\beta(  \\ &
    \qquad \mathbb{D}_{\mathrm{KL}}(q^{\boldsymbol{c}}(\boldsymbol{x}^{w}_{0}|\boldsymbol{x}^{w}_{t},\boldsymbol{x}^{w}_{0}(t))\lVert p^{\boldsymbol{c}}_{\theta}(\boldsymbol{x}^{w}_{0}|\boldsymbol{x}^{w}_{t}))
    -\mathbb{D}_{\mathrm{KL}}(q^{\boldsymbol{c}}(\boldsymbol{x}^{w}_{0}|\boldsymbol{x}^{w}_{t},\boldsymbol{x}^{w}_{0}(t))\lVert p^{\boldsymbol{c}}_{\mathrm{ref}}(\boldsymbol{x}^{w}_{0}|\boldsymbol{x}^{w}_{t}))\\
    & \qquad -\mathbb{D}_{\mathrm{KL}}(q^{\boldsymbol{c}}(\boldsymbol{x}^{l}_{0}|\boldsymbol{x}^{l}_{t},\boldsymbol{x}^{l}_{0}(t))\lVert p^{\boldsymbol{c}}_{\theta}(\boldsymbol{x}^{l}_{0}|\boldsymbol{x}^{l}_{t}))
    +\mathbb{D}_{\mathrm{KL}}(q^{\boldsymbol{c}}(\boldsymbol{x}^{l}_{0}|\boldsymbol{x}^{l}_{t},\boldsymbol{x}^{l}_{0}(t))\lVert p^{\boldsymbol{c}}_{\mathrm{ref}}(\boldsymbol{x}^{l}_{0}|\boldsymbol{x}^{l}_{t})))
\end{aligned}
    \label{eq:suppdpoddim6}
\end{equation}
Using \cref{eq:xtsolution} and the definition of the \textquoteleft initial' variable(\cref{eq:initialvariable}), we can simplify the aforementioned loss function as:
\begin{equation}
\begin{aligned}
    \mathcal{L}(\theta)=& -\mathbb{E}_{(\boldsymbol{x}^{w}_{0},\boldsymbol{x}^{l}_{0}, \boldsymbol{c})\sim \mathcal{D},t \sim \mathcal{U}(0,T), \boldsymbol{x}_{t}^{w} \sim p^{\boldsymbol{c}}_{\theta}(\boldsymbol{x}_{t}^{w}|\boldsymbol{x}_{0}^{w}), \boldsymbol{x}_{t}^{l} \sim p^{\boldsymbol{c}}_{\theta}(\boldsymbol{x}_{t}^{l}|\boldsymbol{x}_{0}^{l})} \log \sigma (-\beta w(t)(\\ 
    &\qquad \lVert \tau^{w}_{t}-\epsilon_{\theta}^{t}(\boldsymbol{x}_{t}^{w},\boldsymbol{c}) \rVert_{2}^{2}-\lVert \tau^{w}_{t}-\epsilon_{\mathrm{ref}}^{t}(\boldsymbol{x}_{t}^{w},\boldsymbol{c}) \rVert_{2}^{2}-\lVert \tau^{l}_{t}-\epsilon_{\theta}^{t}(\boldsymbol{x}_{t}^{l},\boldsymbol{c}) \rVert_{2}^{2}+\lVert \tau^{l}_{t}-\epsilon_{\mathrm{ref}}^{t}(\boldsymbol{x}_{t}^{l},\boldsymbol{c}) \rVert_{2}^{2}))
\end{aligned}
\label{eq:supplossinversionconcrete}
\end{equation}
where $\tau_{t}^{*}= \frac{\boldsymbol{x}_{t}^{*}-\sqrt{\alpha_{t}}\boldsymbol{x}^{*}_{0}}{\sqrt{1-\alpha_{t}}}=\frac{\sqrt{\alpha_{t}}\boldsymbol{x}^{*}_{0}(t)+\sqrt{1-\alpha_{t}}\delta_{t}(\boldsymbol{x}^{*}_{0}(t))-\sqrt{\alpha_{t}}\boldsymbol{x}^{*}_{0}}{\sqrt{1-\alpha_{t}}}  =(\boldsymbol{x}_{0}^{*}(t)-\boldsymbol{x}^{*}_{0})/\sigma_{t}+\delta_{t}(\boldsymbol{x}^{*}_{0}(t))$ and $w(t)$ is a weight function defined consistent with \cite{wallace2024diffusion}.

\paragraph{DDIM ODE on image space} We now provide a comprehensive derivation of \cref{eq:initialddim}, which represents the reparameterization process outlined in \cref{eq:ddim_ode}. This derivation follows the approach presented in \cite{lukoianov2024score}.  To transform the variables from \cref{eq:initialddim} to \cref{eq:initialvariable}, we must differentiate the equation with respect to $t$ and compute $\frac{\mathrm{d}\bar{\boldsymbol{x}}_{t}}{\mathrm{d}t} $. Performing this differentiation directly yields:

\begin{equation}
\frac{\mathrm{d}\boldsymbol{x}_{0}(t)}{\mathrm{d}t} = \frac{\mathrm{d}\bar{\boldsymbol{x}}_{t}}{\mathrm{d}t} - \epsilon_{\theta}^{t}\left( \frac{\boldsymbol{x}_{t}}{\sqrt{\sigma_{t}^2 + 1}}, \boldsymbol{c} \right) \frac{\mathrm{d}\sigma_{t}}{\mathrm{d}t} - \sigma_{t} \frac{\mathrm{d}}{\mathrm{d}t} \epsilon^{t}_{\theta}\left( \frac{\bar{\boldsymbol{x}}_{t}}{\sqrt{\sigma_{t}^2 + 1}}, \boldsymbol{c} \right).
\label{eq:suppsdi1}
\end{equation}
By solving for $ \frac{\mathrm{d}\bar{\boldsymbol{x}}_{t}}{\mathrm{d}t} $ and combining the result with \cref{eq:initialvariable}, we obtain:
\begin{equation}
\begin{aligned}
\epsilon_{\theta}^{t}\left( \frac{\boldsymbol{x}_{t}}{\sqrt{\sigma_{t}^2 + 1}}, \boldsymbol{c} \right) \frac{\mathrm{d}\sigma_{t}}{\mathrm{d}t} &= \frac{\mathrm{d}\boldsymbol{x}_{0}(t)}{\mathrm{d}t}  + \epsilon_{\theta}^{t}\left( \frac{\boldsymbol{x}_{t}}{\sqrt{\sigma_{t}^2 + 1}}, \boldsymbol{c} \right) \frac{\mathrm{d}\sigma_{t}}{\mathrm{d}t} + \sigma_{t} \frac{\mathrm{d}}{\mathrm{d}t} \epsilon^{t}_{\theta}\left( \frac{\bar{\boldsymbol{x}}_{t}}{\sqrt{\sigma_{t}^2 + 1}}, \boldsymbol{c} \right)\\
\frac{\mathrm{d}\boldsymbol{x}_{0}(t)}{\mathrm{d}t} &= -\sigma_{t} \frac{\mathrm{d}}{\mathrm{d}t} \epsilon^{t}_{\theta}\left( \frac{\bar{\boldsymbol{x}}_{t}}{\sqrt{\sigma_{t}^2 + 1}}, \boldsymbol{c} \right) \\
\frac{\mathrm{d}\boldsymbol{x}_{0}(t)}{\mathrm{d}t} &= -\sigma_{t} \frac{\mathrm{d}}{\mathrm{d}t} \epsilon^{t}_{\theta}\left( \frac{\boldsymbol{x}_{0}(t)+\sigma_{t}\delta_{t}(\boldsymbol{x}_{0}(t))}{\sqrt{\sigma_{t}^2 + 1}}, \boldsymbol{c} \right)\\
\frac{\mathrm{d}\boldsymbol{x}_{0}(t)}{\mathrm{d}t} &= -\sigma_{t}\frac{\mathrm{d}}{\mathrm{d}t}\epsilon_{\theta}^{t}\left(\sqrt{\alpha_{t}}\boldsymbol{x}_{0}(t)+\sqrt{1-\alpha_{t}}\delta_{t}(\boldsymbol{x}_{0}(t)),\boldsymbol{c}\right)
\end{aligned}
\label{eq:suppsdi2}
\end{equation}
 It represents a velocity field that maps an initial image to a conditional distribution learned by the diffusion model. The discretized form of inversion leads to \cref{eq:intialddiminversion}.

\suppsection{Choice of $\delta_{t}$} 
\label{sec:suppdelta}
A crucial component of our algorithm is the computation and selection of noise. In theory, 
$\delta_{t}(\boldsymbol{x}_{0}(t))$ should satisfy \cref{eq:nonlinear}. Howeverm, solving it directly is computationally impractical. We present several alternative approaches that are less effective compared to our method:
\begin{itemize}
    \item Fixed-Point Iteration: Given that the model is a known function, we initialize the process with a random noise $\sim \mathcal{N}(0,I)$ as the initial guess and refine the solution iteratively using a predefined iterative formula.
    \item Gradient Optimization Methods: Approaches such as Newton's method or SGD, initialized with random noise $\sim \mathcal{N}(0,I)$, are employed to refine the solution iteratively.
\end{itemize}
These methods for estimating the equation are computationally intensive and lack efficiency. Consequently, we opted not to focus on such equation-solving approaches and instead adopted DDIM inversion for its efficiency and practicality.

\suppsection{Discussion of dataset}
\label{sec:suppcode}
The Pick-a-Pic v2 train and test sets have several notable drawbacks, including the exploitation of vulnerable groups, misrepresentation or defamation of real individuals, and the portrayal of unrealistic or objectifying body imagery. Additionally, it contains harmful or offensive content, as well as explicit or sexual material. These issues emphasize the need for careful curation to avoid ethical concerns and ensure the dataset's responsible use. 

\begin{figure*}[t]
  \centering
   \includegraphics[width=0.55\linewidth]{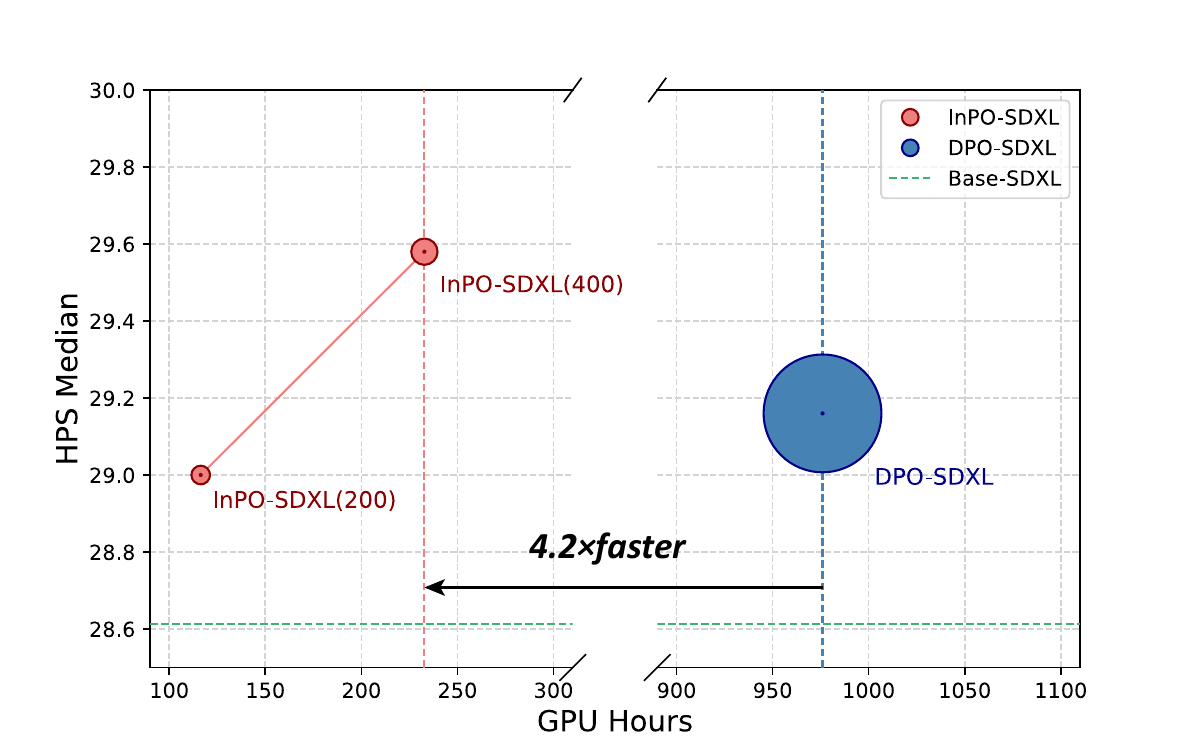}

   \caption{Comparison of the trade-off between the quality of generated images and training efficiency following human preference optimization of SDXL on the HPDv2 test set. Sizes of the circles represent the volume of training data used. Our DDIM-InPO achieves superior performance, with a training speed that is 4.2 times faster than Diffusion-DPO while producing images of higher quality. }
   \label{fig:suppsdxlspeed}
\end{figure*}

\begin{figure*}[ht]
  \centering
   \includegraphics[width=0.55\linewidth]{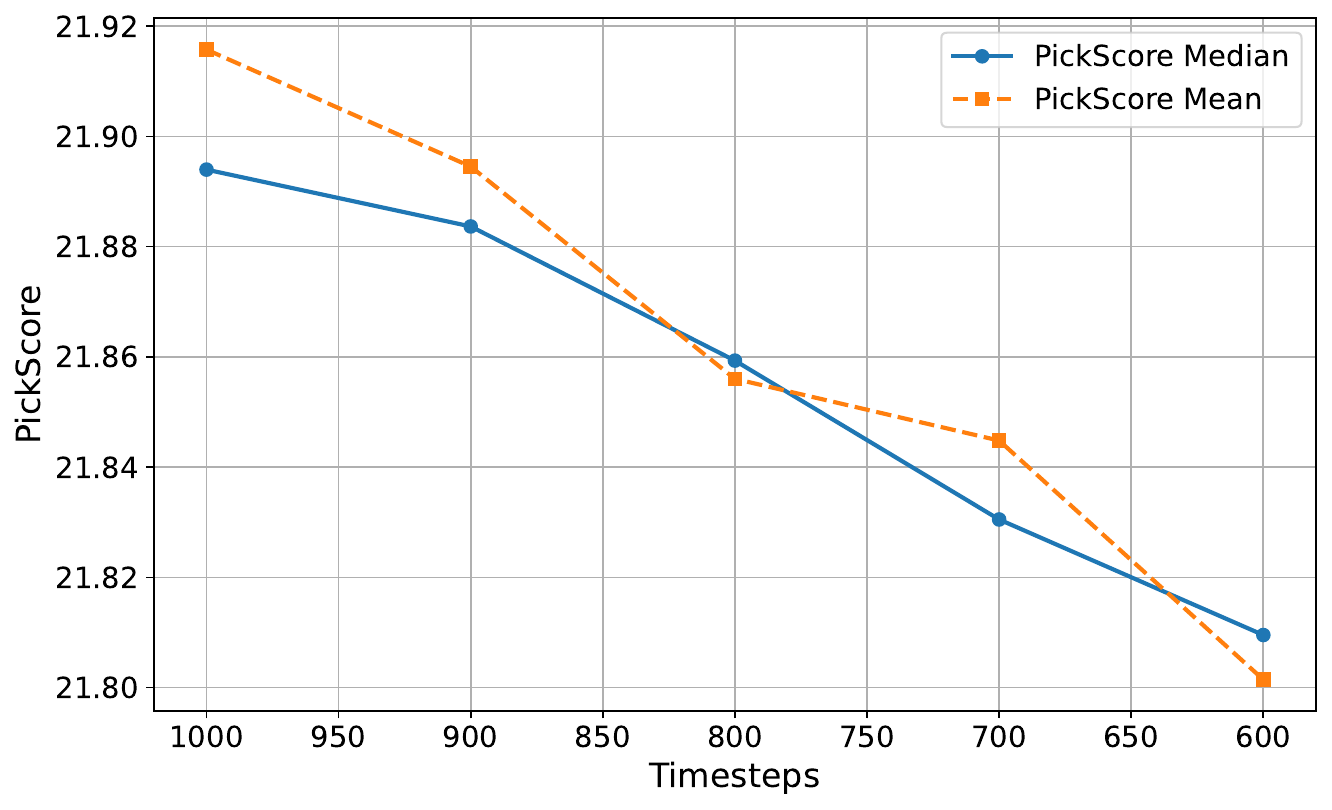}

   \caption{Timestep ablation studies of our DDIM-InPO method for fine-tuning SD1.5, evaluated on the HPDv2 test set using both median and mean PickScore metrics.}
   \label{fig:suppablation}
\end{figure*}

\suppsection{Further Discussion}
\label{sec:suppdiscussion}
In T2I diffusion models, human preference feedback is shaped by factors such as image quality, realism, artistic style, and cultural background. These factors are highly subjective, and the presence of noise in datasets makes it challenging for AI to effectively learn from them, underscoring the importance of robust preference learning. Additionally, the diversity and inherent uncertainty of human preferences during the T2I diffusion process introduce significant modeling complexities and may lead to distributional shifts. For instance, most preference data is derived from Stable Diffusion variants, and applying this data to other T2I models (e.g., Midjourney or DALL·E 3) may result in distributional mismatches, causing inconsistencies between the model outputs and human preferences. This issue arises because these models are trained on distinct data distributions, leading to potential training conflicts. Furthermore, feedback optimized for one model might fail to capture the nuanced preferences required by another, further compounding the problem. As a result, preference alignment techniques must carefully address these challenges to ensure consistent and robust performance across different T2I diffusion models. We propose that supervised fine-tuning, particularly when dataset quality surpasses that of the model, can be employed to reduce the impact of distributional shifts. Nonetheless, we reaffirm that state-of-the-art results can still be achieved without supervised fine-tuning using our method, as shown in \cref{tab:winrate} and \cref{tab:ablation}.
\suppsection{Experiment Details} 
\label{sec:suppexperimentdetals}
\paragraph{Pick-a-Pic v2}
The Pick-a-Pic dataset is a text-to-image pair dataset that gathers user feedback from the Pick-a-Pic web application. Each image pair (comprising two images) is associated with a text prompt and a label reflecting the user's preference. The dataset includes images generated by various text-to-image models, such as Stable Diffusion 2.1, Dreamlike Photoreal, and variants of Stable Diffusion XL, with a wide range of Classifier-Free Guidance (CFG) values. In this paper, we use its training data. In the supplementary materials, we provided additional quantitative comparative analyses on the test dataset to further validate our approach.
\paragraph{HPDv2} 
HPDv2 collects human preference data via the "Dreambot" channel on the Stable Foundation Discord server. It contains 25,205 text prompts used to generate a total of 98,807 images. Each text prompt is associated with multiple generated images and paired image labels, where the label denotes the user's preferred choice between two images. The number of generated images per text prompt varies across the dataset. In this paper, we use its test data of text prompt (3200 prompts).
\paragraph{Parti-Prompts} 
Parti-Prompts is a comprehensive dataset consisting of 1,632 text prompts, specifically designed to evaluate and benchmark the capabilities of text-to-image generation models. Covering multiple categories, these prompts offer diverse challenges that facilitate a thorough assessment of model performance across various dimensions.
\paragraph{Additional Implementation details}
 During the evaluation and comparison phase for text-to-image generation, the inference CFG is set to 7.5 for SD1.5 and 5 for SDXL, which are widely recognized as standard and recommended configurations. In generation tasks conditioned on depth maps and canny edges, we set the ControlNet conditioning scale to 0.5 and the CFG to 5. For inpainting tasks, the strength parameter is set to 0.85, with the CFG also set to 5.
\paragraph{Chart Explanation} 
To facilitate clearer comparisons, we scale PickScore, HPS and the CLIP score by a factor of 100 and retain 5 significant figures for precision, including \cref{fig:speedsd15}, \cref{fig:rlaif}, \cref{fig:suppsdxlspeed}, \cref{fig:suppablation}, \cref{tab:ablation}, \cref{tab:suppsdxlhpdv2}, \cref{tab:suppsd15hpdv2}, \cref{tab:suppsdxlpartiprompts},  \cref{tab:suppsd15partiprompts}, \cref{tab:suppsdxlpicktest}, \cref{tab:suppsd15picktest}.

\suppsection{Additional Quantitative Results}
In this section, we present additional qualitative results. First, we present additional comparison of the trade-off between the quality of generated images and training efficiency of SDXL on the HPDv2 test set. Subsequently, we also provide automatic preference evaluation comparison of different prompt sets conducted on SDXL and SD1.5. 
\begin{itemize}
    \item \cref{fig:suppsdxlspeed} shows the comparison of the trade-off between the quality of generated images and training efficiency following human preference optimization of SDXL on the HPDv2 test set. We present the results of training for 200 and 400 steps using our DDIM-InPO and compare them with those of Diffusion-DPO finetuned for 2000 steps. 
    \item \cref{tab:suppsdxlhpdv2} and \cref{tab:suppsd15hpdv2} demonstrate evaluation comparison on SDXL and SD1.5 using  HPDv2 test set, respectively.
    \item \cref{tab:suppsdxlpartiprompts} and \cref{tab:suppsd15partiprompts} demonstrate evaluation comparison on SDXL and SD1.5 using Parti-Prompts, respectively. 
    \item \cref{tab:suppsdxlpicktest} and \cref{tab:suppsd15picktest} demonstrate evaluation comparison on SDXL and SD1.5 using Pick-a-Pic test set., respectively.
\end{itemize}
\paragraph{Experimental Result Analysis}
Overall, following fine-tuning with DDIM-InPO, both SD1.5 and SDXL achieve superior performance compared to the baselines across nearly all evaluators and test datasets, thereby validating the effectiveness of our method. \cref{fig:suppsdxlspeed} shows that our DDIM-InPO achieves better performance, with a training speed that is 4.2 times faster than Diffusion-DPO while producing images of higher quality. All tables clearly show that while supervised fine-tuning performs well on SD1.5, its application to SDXL results in significant degradation of the base model, making it an ineffective and non-generalizable approach. This limitation stems from the dataset quality, as the Pick-a-Pic dataset is inferior to the outputs generated by the SDXL base model. In comparison, Diffusion-DPO proves to be a more robust alternative, delivering consistent improvements on both SD1.5 and SDXL. However, although Diffusion-KTO achieves notable gains on SD1.5, its high computational demands prevent effective scalability to SDXL models. By contrast, our model emerges as a more effective and efficient solution, achieving state-of-the-art results across nearly all evaluators and test datasets on both SDXL and SD1.5. These findings highlight the suitability of our approach for diffusion models, along with its significant advantage in training speed.

\newpage
\label{sec:suppquanti}
\begin{table*}[t]
    \centering
    \begin{tabular}{lcccccccc}
        \toprule
          \multirow{2}{*}{Baselines}& \multicolumn{2}{c}{Aesthetic} & \multicolumn{2}{c}{PickScore}& \multicolumn{2}{c}{HPS} & \multicolumn{2}{c}{CLIP} \\
         & Median  & Mean & Median  & Mean  & Median  & Mean & Median  & Mean \\
         \midrule
         Base-SDXL& \underline{6.1143} & \underline{6.1346} & 22.756 & 22.781 & 28.614  & 28.624 & 38.360 & 38.155  \\
          SFT-SDXL&  5.8049 & 5.8327 & 21.524 & 21.380 & 27.467  & 27.204 & 37.217 & 36.531 \\
          DPO-SDXL& 6.1124 & 6.1310 & \underline{23.133} & \underline{23.152} & \underline{29.165}  & \underline{29.174} & \textbf{38.865} & \textbf{38.711} \\
          InPO-SDXL& \textbf{6.1676} & \textbf{6.1820} & \textbf{23.254} & \textbf{23.274} &\textbf{29.576}  & \textbf{29.550} & \underline{38.627} & \underline{38.449}  \\
         \bottomrule 
    \end{tabular}
    \caption{Automatic preference evaluation comparison to existing alignment baselines on SDXL using prompts from HPDv2 test set. We use median and mean values of four evaluators. To ensure clarity in comparisons, Pickscore, HPS, and CLIP scores are scaled by 100, and all evaluator values retain precision to five significant figures. In the table, the maximum value in each column is bolded, while the second-highest value is underlined.}
    \label{tab:suppsdxlhpdv2}
\end{table*}

\begin{table*}[t]
    \centering
    \begin{tabular}{lcccccccc}
        \toprule
          \multirow{2}{*}{Baselines}& \multicolumn{2}{c}{Aesthetic} & \multicolumn{2}{c}{PickScore}& \multicolumn{2}{c}{HPS} & \multicolumn{2}{c}{CLIP} \\
         & Median  & Mean & Median  & Mean  & Median  & Mean & Median  & Mean \\
         \midrule
         Base-SD1.5& 5.3491 & 5.3848 & 20.719 & 20.727 & 26.647  & 26.633 & 34.276 & 33.945  \\
          SFT-SD1.5& \underline{5.7255} & \underline{5.7515} & \underline{21.647} & \underline{21.648} & 28.032  & 27.977 & \underline{36.292} &  \underline{35.845} \\
          DPO-SD1.5& 5.5219 & 5.5841 & 21.274 & 21.297 & 27.428  & 27.392 & 35.591 & 35.197 \\
          KTO-SD1.5& 5.6922 & 5.7248 & 21.566 & 21.582 & \underline{28.376}  & \underline{28.306} & 35.902 & 35.648  \\
          InPO-SD1.5& \textbf{5.7734} & \textbf{5.8056} & \textbf{21.894} & \textbf{21.916} &\textbf{28.523}  & \textbf{28.502} & \textbf{36.876} & \textbf{36.495}  \\
         \bottomrule 
    \end{tabular}
    \caption{Automatic preference evaluation comparison to existing alignment baselines on SD1.5 using prompts from HPDv2 test set. We use median and mean values of four evaluators. To ensure clarity in comparisons, Pickscore, HPS, and CLIP scores are scaled by 100, and all evaluator values retain precision to five significant figures. In the table, the maximum value in each column is bolded, while the second-highest value is underlined.}
    \label{tab:suppsd15hpdv2}
\end{table*}

\begin{table*}[t]
    \centering
    \begin{tabular}{lcccccccc}
        \toprule
          \multirow{2}{*}{Baselines}& \multicolumn{2}{c}{Aesthetic} & \multicolumn{2}{c}{PickScore}& \multicolumn{2}{c}{HPS} & \multicolumn{2}{c}{CLIP} \\
         & Median  & Mean & Median  & Mean  & Median  & Mean & Median  & Mean \\
         \midrule
         Base-SDXL& 5.7519 & 5.7681 & 22.648 & 22.628 & 28.447  & 28.424 & 35.550 & 35.531  \\
          SFT-SDXL& 5.5373  & 5.5403 & 21.666 & 21.554 & 27.213  & 27.085 & 34.827 &  34.696 \\
          DPO-SDXL& \underline{5.8181} & \underline{5.7942} & \underline{22.91}4 & \underline{22.928} & \underline{28.885}  & \underline{28.906} & \textbf{36.401} & \textbf{36.457} \\
          InPO-SDXL& \textbf{5.8493} & \textbf{5.8566} & \textbf{23.039} & \textbf{23.005} &\textbf{29.123}  & \textbf{29.143} & \underline{35.914} & \underline{35.903}  \\
         \bottomrule 
    \end{tabular}
    \caption{Automatic preference evaluation comparison to existing alignment baselines on SDXL using prompts from Parti-Prompts. We use median and mean values of four evaluators. To ensure clarity in comparisons, Pickscore, HPS, and CLIP scores are scaled by 100, and all evaluator values retain precision to five significant figures. In the table, the maximum value in each column is bolded, while the second-highest value is underlined.}
    \label{tab:suppsdxlpartiprompts}
\end{table*}

\begin{table*}[t]
    \centering
    \begin{tabular}{lcccccccc}
        \toprule
          \multirow{2}{*}{Baselines}& \multicolumn{2}{c}{Aesthetic} & \multicolumn{2}{c}{PickScore}& \multicolumn{2}{c}{HPS} & \multicolumn{2}{c}{CLIP} \\
         & Median  & Mean & Median  & Mean  & Median  & Mean & Median  & Mean \\
         \midrule
         Base-SD1.5& 5.3494 & 5.3132 & 21.406 & 21.389 & 27.291  & 27.172 & 33.065 & 33.128  \\
          SFT-SD1.5& \underline{5.5798}  & \underline{5.5506}  & \underline{21.803} & \underline{21.759} & 28.192 & \underline{28.129}  & 33.887 & 33.956  \\
          DPO-SD1.5& 5.4445 & 5.3874 & 21.619 & 21.631 & 27.596  & 27.511 & 33.551 & 33.694 \\
          KTO-SD1.5& 5.5466 & 5.5110 & 21.755 & 21.736 & \underline{28.240}  & 28.110 & \underline{34.101} & \underline{34.013}  \\
          InPO-SD1.5& \textbf{5.6056} & \textbf{5.5698} & \textbf{21.957} & \textbf{21.923}   & \textbf{28.431} & \textbf{28.325} & \textbf{34.533} &\textbf{34.683} \\
         \bottomrule 
    \end{tabular}
    \caption{Automatic preference evaluation comparison to existing alignment baselines on SD1.5 using prompts from Parti-Prompts. We use median and mean values of four evaluators. To ensure clarity in comparisons, Pickscore, HPS, and CLIP scores are scaled by 100, and all evaluator values retain precision to five significant figures. In the table, the maximum value in each column is bolded, while the second-highest value is underlined.}
    \label{tab:suppsd15partiprompts}
\end{table*}

\begin{table*}[t]
    \centering
    \begin{tabular}{lcccccccc}
        \toprule
          \multirow{2}{*}{Baselines}& \multicolumn{2}{c}{Aesthetic} & \multicolumn{2}{c}{PickScore}& \multicolumn{2}{c}{HPS} & \multicolumn{2}{c}{CLIP} \\
         & Median  & Mean & Median  & Mean  & Median  & Mean & Median  & Mean \\
         \midrule
         Base-SDXL& 5.9775 & 6.0057 & 22.219 & 22.159 & 27.995  & 27.978 & 36.470 & 36.124  \\
          SFT-SDXL& 5.6416 & 5.6452 & 21.028 & 21.003 & 26.790 & 26.681 & 35.589 &  35.427 \\
          DPO-SDXL& \underline{6.0179} & \underline{6.0160} & \underline{22.581} & \underline{22.627} & \underline{28.515} & \underline{28.586} & \textbf{37.404} & \textbf{37.392} \\
          InPO-SDXL& \textbf{6.0372} & \textbf{6.0558} & \textbf{22.606} & \textbf{22.692} &\textbf{28.824}  & \textbf{28.817} & \underline{37.130} & \underline{36.842}  \\
         \bottomrule 
    \end{tabular}
    \caption{Automatic preference evaluation comparison to existing alignment baselines on SDXL using prompts from Pick-a-Pic v2 test set. We use median and mean values of four evaluators. To ensure clarity in comparisons, Pickscore, HPS, and CLIP scores are scaled by 100, and all evaluator values retain precision to five significant figures. In the table, the maximum value in each column is bolded, while the second-highest value is underlined.}
    \label{tab:suppsdxlpicktest}
\end{table*}

\begin{table*}[t]
    \centering
    \begin{tabular}{lcccccccc}
        \toprule
          \multirow{2}{*}{Baselines}& \multicolumn{2}{c}{Aesthetic} & \multicolumn{2}{c}{PickScore}& \multicolumn{2}{c}{HPS} & \multicolumn{2}{c}{CLIP} \\
         & Median  & Mean & Median  & Mean  & Median  & Mean & Median  & Mean \\
         \midrule
         Base-SD1.5& 5.3545 & 5.3296 & 20.632 & 20.661 & 26.527  & 26.480 & 33.023 & 32.619  \\
          SFT-SD1.5& \underline{5.6441} & \underline{5.6285} & \underline{21.278} & \underline{21.253} & \underline{27.707} & 27.509 & \underline{34.050} & \underline{34.144}  \\
          DPO-SD1.5& 5.5258 & 5.4654 & 21.020 & 21.053 & 27.098  & 26.913 & 33.270 & 33.302 \\
          KTO-SD1.5& 5.6029 & 5.5831 & 21.184 & 21.190 & 27.645  & \underline{27.580} & 34.003 & 33.910  \\
          InPO-SD1.5& \textbf{5.6810} & \textbf{5.6585} & \textbf{21.456} & \textbf{21.490}   & \textbf{27.866} & \textbf{27.765} & \textbf{34.782} &\textbf{34.728} \\
         \bottomrule 
    \end{tabular}
    \caption{Automatic preference evaluation comparison to existing alignment baselines on SD1.5 using prompts from Pick-a-Pic v2 test set. We use median and mean values of four evaluators. To ensure clarity in comparisons, Pickscore, HPS, and CLIP scores are scaled by 100, and all evaluator values retain precision to five significant figures. In the table, the maximum value in each column is bolded, while the second-highest value is underlined.}
    \label{tab:suppsd15picktest}
\end{table*}

\suppsection{Additional AI Preference}
\label{sec:supprlaif}
\begin{figure*}[ht]
  \centering
   \includegraphics[width=0.6\linewidth]{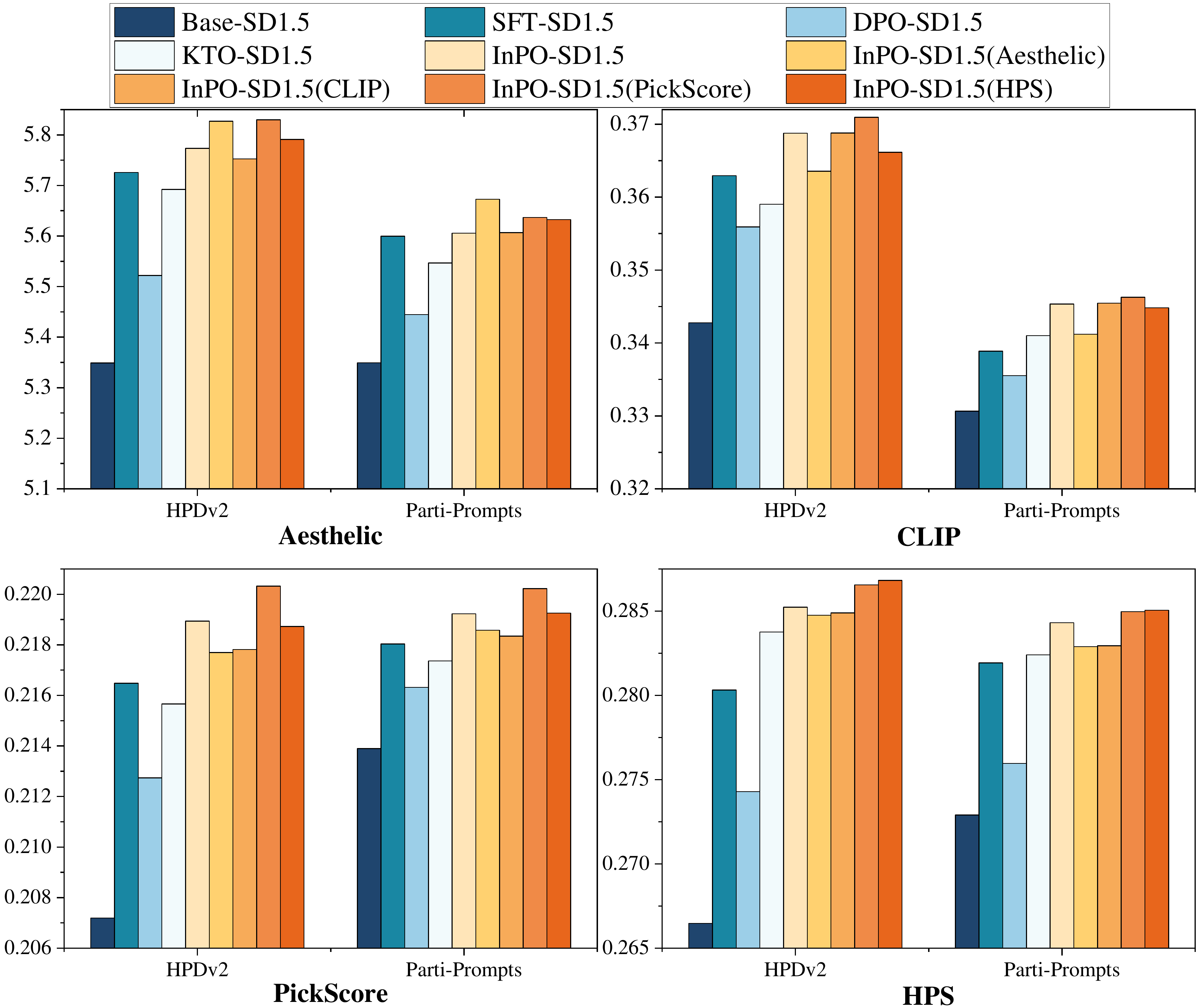}

   \caption{Median of Aesthelc, CLIP score, PickScore and HPS comparisons for all baselines and test datasets on SD1.5.}
   \label{fig:supprlaif}
\end{figure*}
In this section, we extend the AI preference experiments discussed in the main text. Additionally, we incorporate aesthetic classifiers and CLIP as evaluators, where higher scores reflect stronger AI preferences. \cref{fig:supprlaif} reveal that training with self-selected images leads to improved scores. Specifically, training with images selected by the aesthetic classifier results in higher aesthetic metrics, and a similar pattern is observed for Pickscore and HPS. Our findings indicate that Pickscore and HPS effectively emulate human preferences, enabling models trained with these metrics to surpass InPO-SD1.5. Conversely, models trained with CLIP-based preference selection exhibit relatively lower scores, suggesting that text alignment plays a less significant role in preference selection.
\suppsection{Additional Ablations}
\label{sec:suppablation}
In this section, we introduce additional ablation experiments, primarily exploring whether the number of training timesteps can be reduced. Specifically, we consider the denoiser as timestep-aware, with the total timesteps for DDPM denoising set to 1000. Can we train only on the last 900, 800, or even fewer timesteps to accelerate the training process?

\noindent As shown in \cref{fig:suppablation}, our method demonstrates strong robustness. While reducing the training timesteps results in a slight performance decline, it still achieves significant improvements over baselines such as KTO-SD1.5, DPO-SD1.5, and SFT-SD1.5 (refer to \cref{tab:suppsd15hpdv2}). This suggests potential for future exploration in optimizing training efficiency by reducing timesteps. Furthermore, investigating the trade-off between training timesteps and inversion steps could provide deeper insights into balancing efficiency and performance, further underscoring the potential of our approach.

\begin{figure*}[t]
  \centering
   \includegraphics[width=0.8\linewidth]{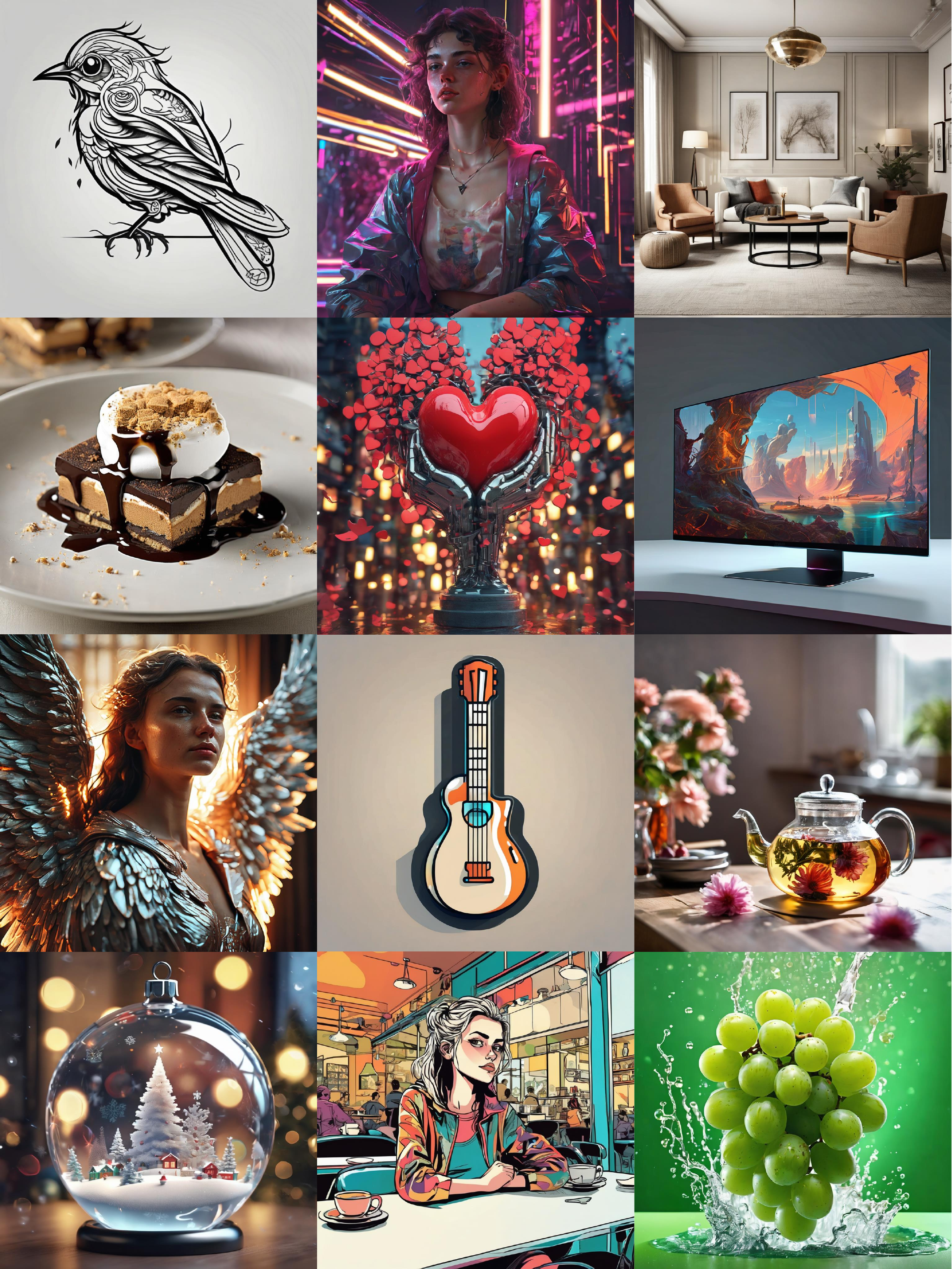}
    
   \caption{\footnotesize Additional qualitative results of InPO-SDXL for \textbf{various prompts}, arranged from left to right and top to bottom. \textit{Prompts: (1) Tattoo design: a tattoo design, a small bird, minimalistic, black and white drawing, detailed, 8k. (2) Young glitchy woman, beautiful girl, 8k, unreal engine, illustration, trending on artstation, masterpiece. (3) A three-seater sofa, capet,end table,west elm chandelier,armchair, modrtn organic style, crrisp lines, neutral colors, backdrop of simplicity, bright environment. (4) An indulgent dessert featuring charred marshmallow, chocolate fondant, and graham cracker crumbs. (5) Hearts, in the style of jamie hewlett killian eng kawase hasui riyoko ikeda, artstation trending, 8 k, octane render, photorealistic, volumetric lighting caustics, surreal. (6) A sleek, ultra-thin, high resolution bezel-less monitor mockup, realistic, modern, digital illustration, trending on Artstation, high-tech, smooth, minimalist workstation background, crisp reflection on screen, soft lighting. (7) Portrait art of female angel, art by alessio albi 8 k ultra realistic, angel wings, lens flare, atmosphere, glow, detailed, intricate, cinematic lighting, trending on artstation, 4k, hyperrealistic, focused, extreme details, unreal engine 5, masterpiece. (8) Icon: a guitar, 2d minimalistic icon, flat vector illustration, digital, smooth shadows, design asset. (9) Drink photography: freshly made hot floral tea in glass kettle on the table, angled shot, midday warm, Nikon D850 105mm, close-up. (10) Looking through a transparent glass Christmas ball , hyper-realistic, minimalist, futuristic background with cute Christmas decorations like Santa Claus and snowflakes, 8k. (11) Comicbook: a girl sitting in the cafe, comic, graphic illustration, comic art, graphic novel art, vibrant, highly detailed, colored, 2d minimalistic. (12) Powerful liquid explosion, green grapes, green background, commercial photography, a bright environment, studio lighting, OC rendering, isolated platform, professional photography.}}
   \label{fig:suppinpo1}
\end{figure*}

\begin{figure*}[t]
  \centering
   \includegraphics[width=0.9\linewidth]{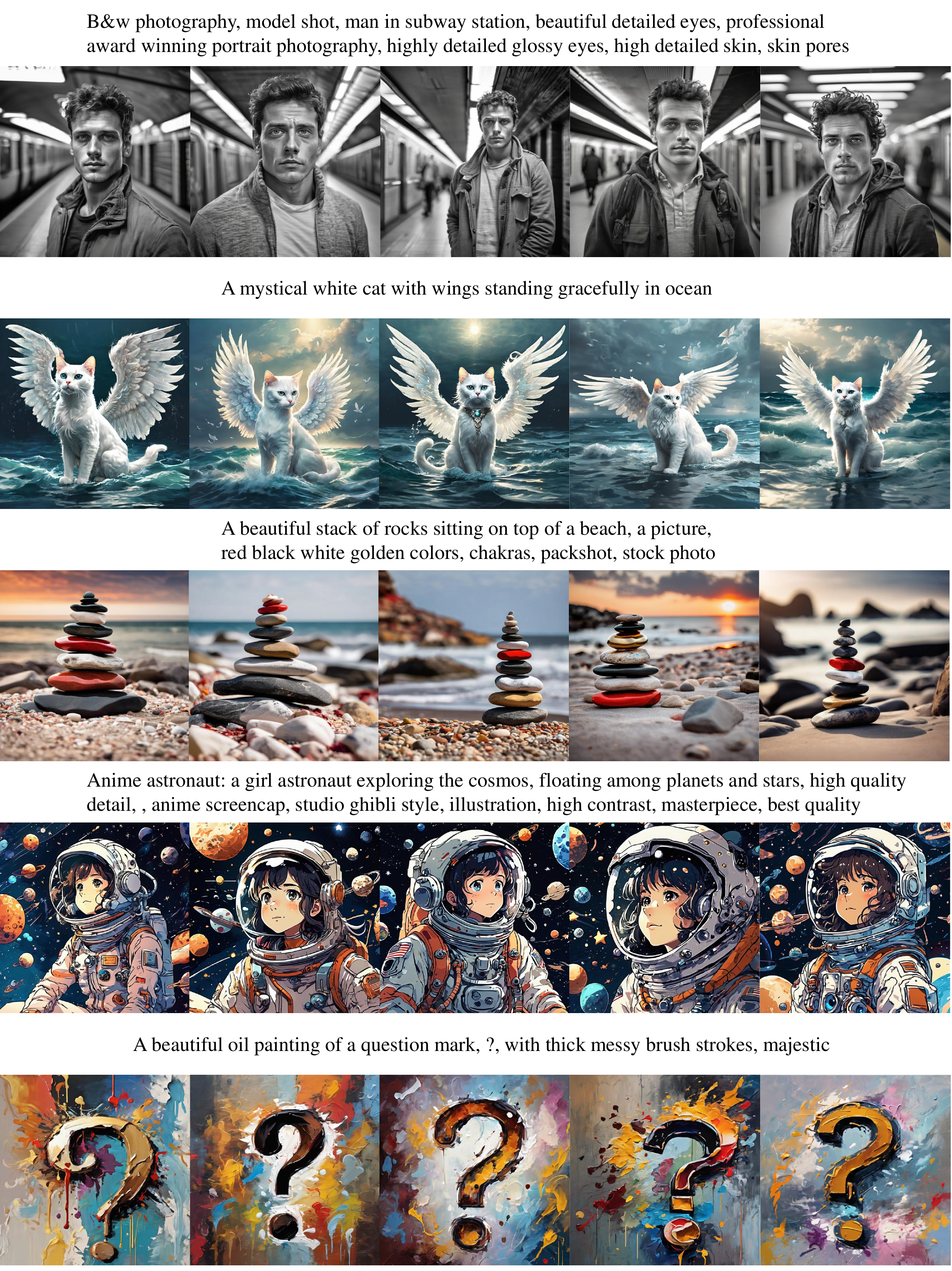}

   \caption{Additional qualitative results of InPO-SDXL (A). The seeds are chosen from the range of 0 to 15. After fine-tuning with our method, the model not only retains its original generative capabilities but also produces images that align with human preferences.}
   \label{fig:suppinpo2}
\end{figure*}

\begin{figure*}[t]
  \centering
   \includegraphics[width=0.9\linewidth]{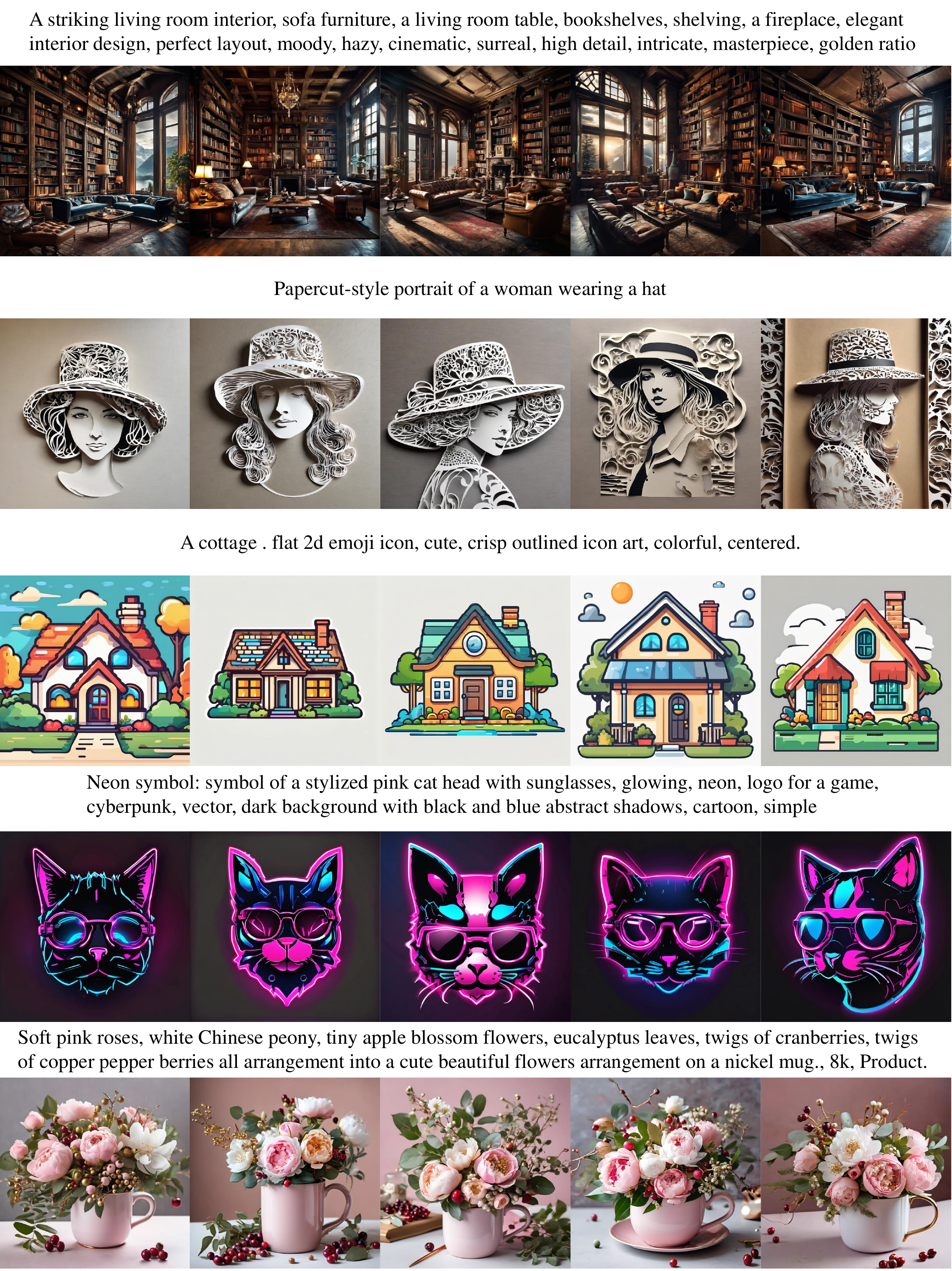}

   \caption{Additional qualitative results of InPO-SDXL (B). The seeds are chosen from the range of 0 to 15. After fine-tuning with our method, the model not only retains its original generative capabilities but also produces images that align with human preferences.}
   \label{fig:suppinpo3}
\end{figure*}

\begin{figure*}[t]
  \centering
   \includegraphics[width=1\linewidth]{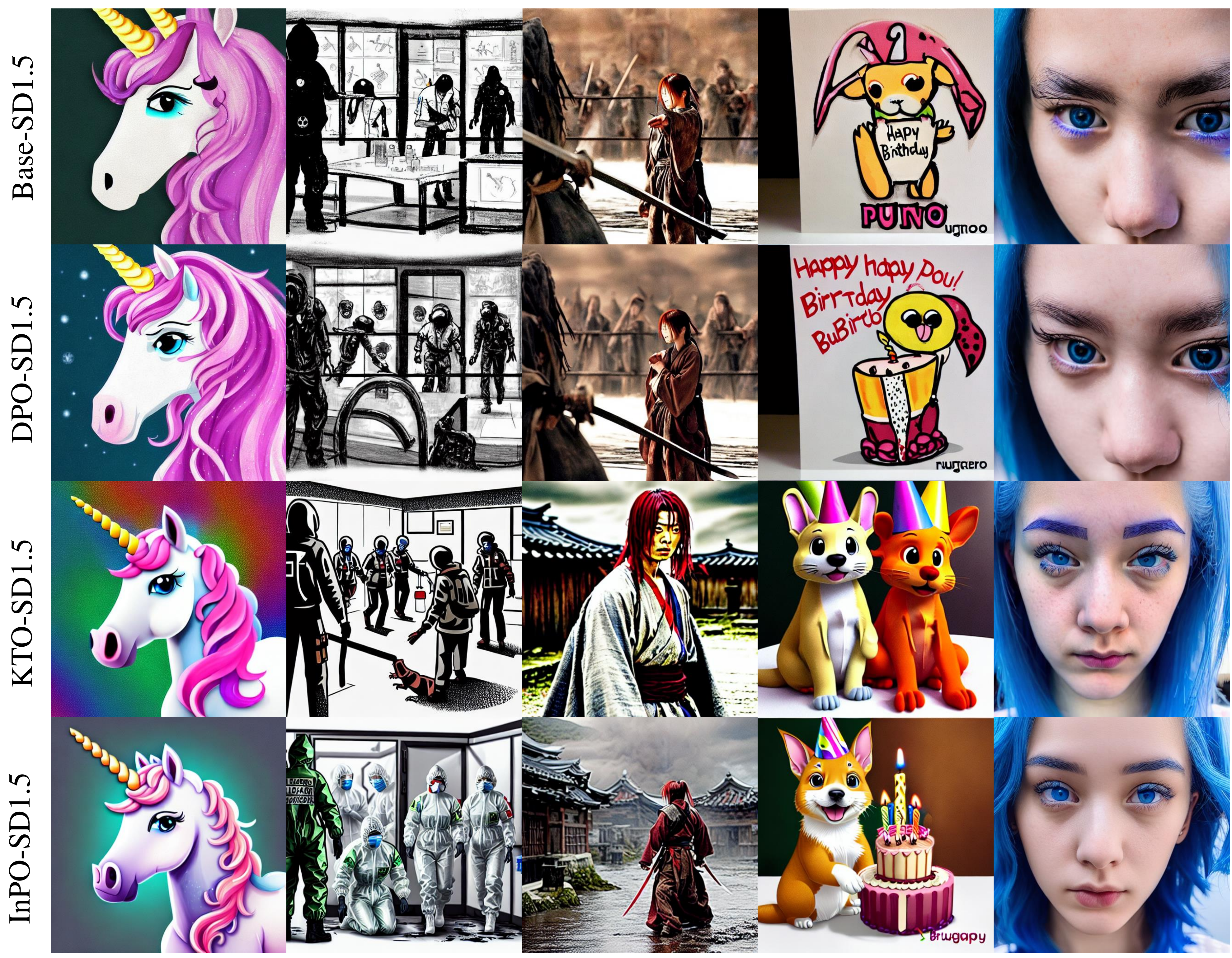}

   \caption{Qualitative comparisons among baselines of SD1.5 (A). InPO-SD1.5 achieves superior prompt alignment and produces images of higher quality.  Prompts from left to right: \textit{(1) A cute digital art of a unicorn. (2) A detailed, realistic image of a biohazard lab evacuation with horror influences and multiple art styles incorporated. (3) An apocalyptic scene from Kenshin. (4) A birthday greeting for Pungeroo. (5) A blue-haired girl with soft features stares directly at the camera in an extreme close-up Instagram picture.}}
   \label{fig:suppsd15_1}
\end{figure*}

\begin{figure*}[t]
  \centering
   \includegraphics[width=1\linewidth]{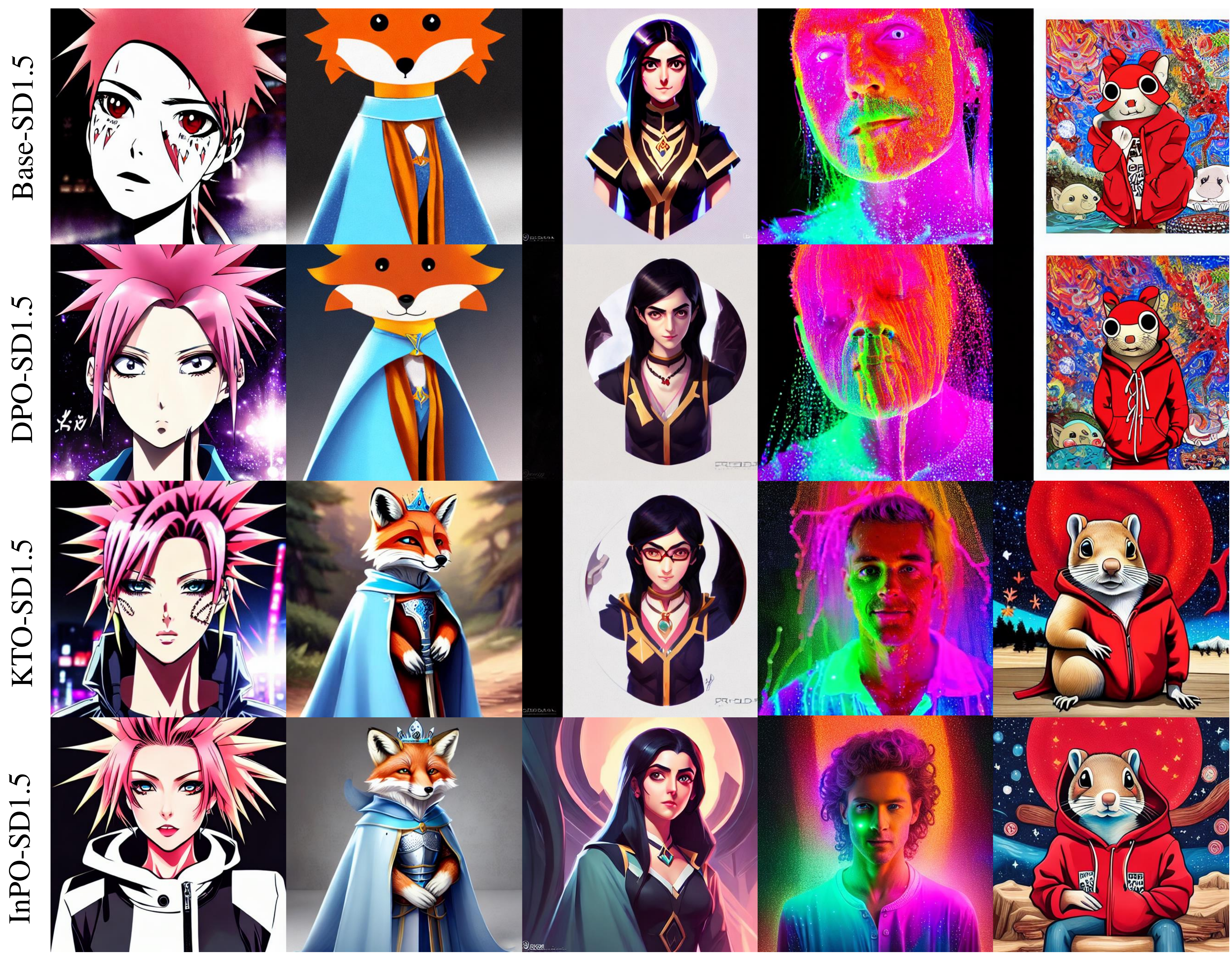}

   \caption{Qualitative comparisons among baselines of SD1.5 (B). InPO-SD1.5 achieves superior prompt alignment and produces images of higher quality.  Prompts from left to right: \textit{(1) Anime poster of a woman wearing futuristic streetwear with spiky hair, featuring intricate eyes and a pretty face. (2) A cute anthropomorphic fox knight wearing a cape and crown in pale blue armor. (3) Head-on centered portrait of Maya Ali as a black-haired RPG mage, depicted in stylized concept art for a Blizzard game, by Lois Van Baarle, Ilya Kuvshinov, and RossDraws. (4) A human portrait formed out of neon rain on a galactic background. (5) A water squirrel spirit wearing a red hoodie sits under the stars, surrounded by artwork from various artists.}}
   \label{fig:suppsd15_2}
\end{figure*}

\begin{figure*}[t]
  \centering
   \includegraphics[width=0.7\linewidth]{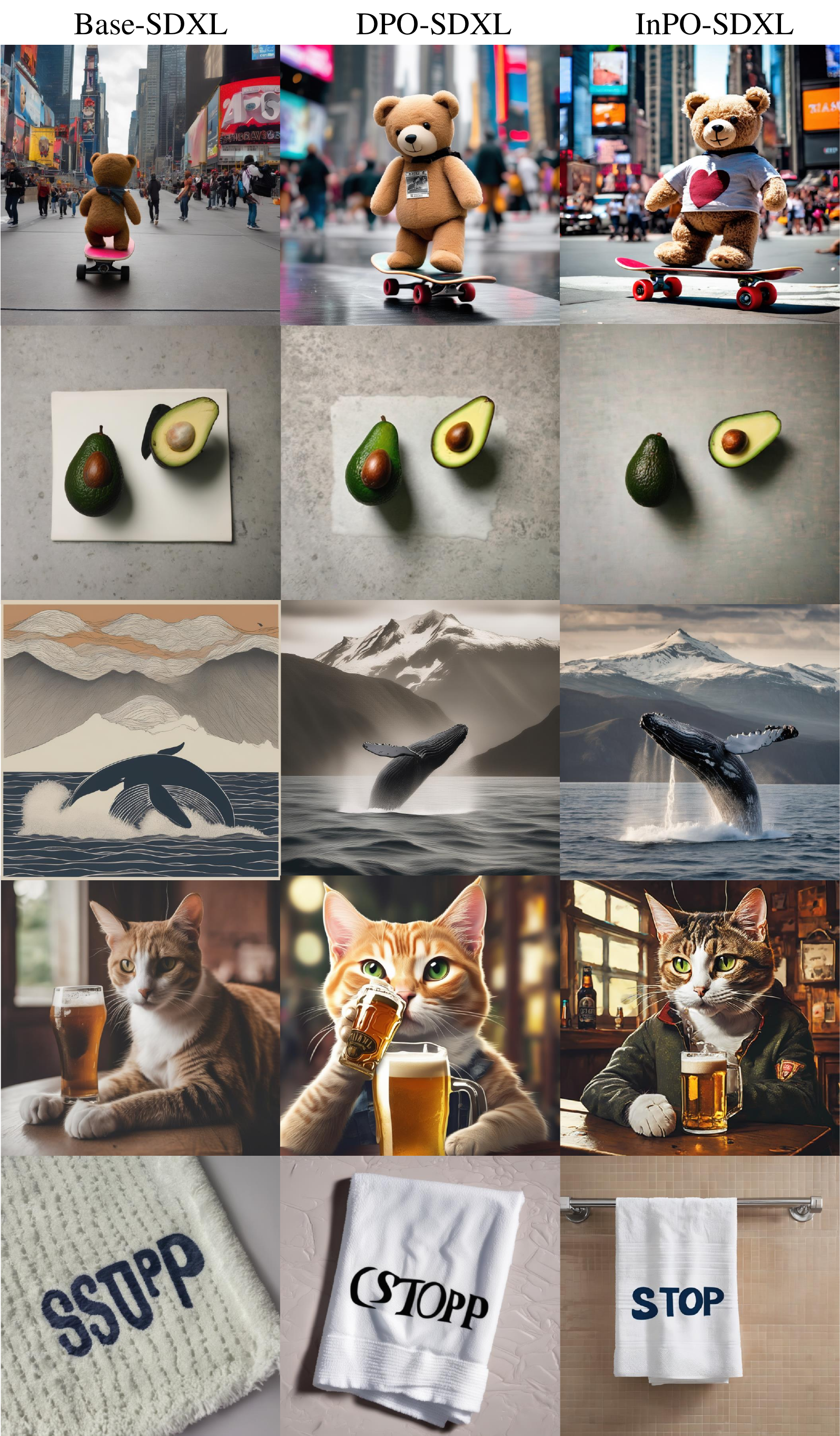}

   \caption{Additional qualitative evaluation of InPO-SDXL in comparison with Base-SDXL and DPO-SDXL on T2I generation tasks. Prompts from top to bottom: \textit{(1) A teddy bear on a skateboard in times square. (2) An avocado on a table. (3) A whale breaching near a mountain. (4)  A cat drinking a pint of beer. (5) A towel with the word \textquoteleft stop’ printed on it, simple and clear text.}}
   \label{fig:suppsdxlcompare}
\end{figure*}

\begin{figure*}[t]
  \centering
   \includegraphics[width=0.85\linewidth]{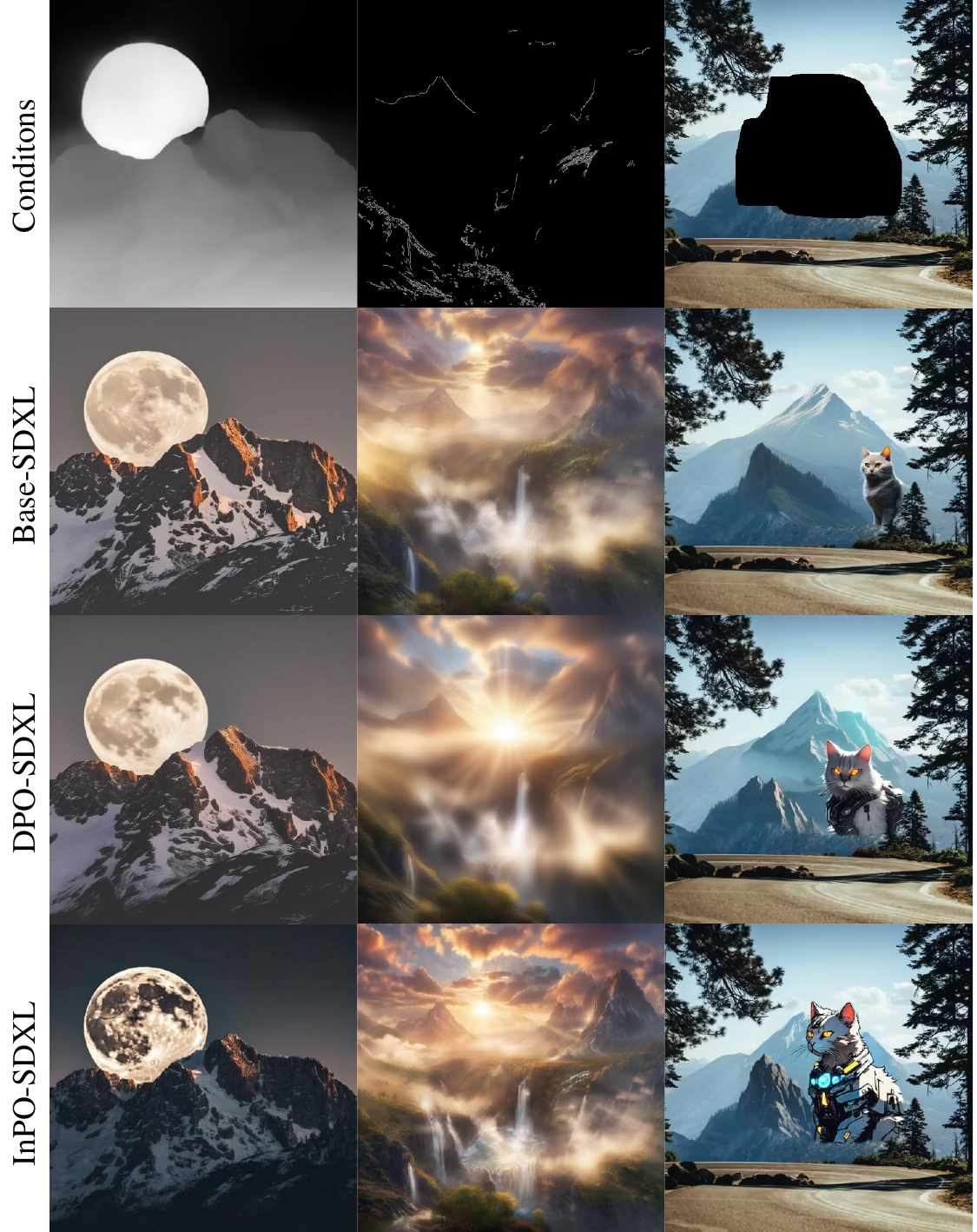}

   \caption{Additional qualitative evaluation of InPO-SDXL in comparison with Base-SDXL and DPO-SDXL on conditional generation tasks (From left to right: depth map, canny edge, and inpainting). Prompts from left to right: \textit{(1) A full moon rising above a mountain at night. (2) Sunset over misty mountains, cascading waterfalls, and soft god rays breaking through clouds, creating a realistic and serene atmosphere. (3) A big cyberpunk cat with glowing eyes, sitting majestically against a mountain backdrop.}}
   \label{fig:suppcondition}
\end{figure*}

\suppsection{Additional Qualitative results}
\label{sec:suppquanli}
In this section, we present additional qualitative results, including evaluations conducted on SD1.5 and SDXL.
\\
\begin{itemize}
    \item \cref{fig:suppinpo1} showcases the qualitative generation results of InPO-SDXL across diverse prompts. \cref{fig:suppinpo2} and \cref{fig:suppinpo3} display outputs generated with different random seeds under the same prompt, where the seeds are randomly chosen from the range 0 to 15. These results highlight that our fine-tuned model not only preserves the capabilities of SDXL but also produces high-quality outputs that align with human preferences.
    \item \cref{fig:suppsd15_1} and \cref{fig:suppsd15_2} provide a qualitative comparison between InPO-SD1.5 and the SD1.5 baselines, using prompts sourced from HPDv2. The results reveal that our model exhibits superior text alignment, enhanced visual appeal, and a stronger consistency with human preferences.
    \item \cref{fig:suppsdxlcompare} provides an additional qualitative qualitative evaluation of InPO-SDXL in comparison with Base-SDXL and DPO-SDXL on T2I generation tasks, further highlighting the effectiveness of our approach in demonstrating improved performance.
    \item \cref{fig:suppcondition} provides an additional qualitative evaluation of InPO-SDXL in comparison with Base-SDXL and DPO-SDXL on conditional generation tasks, including depth map, canny edge, and inpainting. 
    \item Due to safety concerns, failed cases are unsuitable for presentation in the paper. A small subset of images occasionally exhibit an excessively feminized style.
\end{itemize}

\end{document}